\documentclass[10pt,twocolumn,letterpaper]{article}

\usepackage{cvpr}              

\definecolor{cvprblue}{rgb}{0.21,0.49,0.74}
\usepackage[pagebackref,breaklinks,colorlinks,allcolors=cvprblue]{hyperref}

\usepackage{mathtools}

\usepackage{amsmath,amsfonts,bm}

\def\eqref#1{equation~\ref{#1}}

\def\1{\bm{1}}

\DeclareMathAlphabet{\mathsfit}{\encodingdefault}{\sfdefault}{m}{sl}
\SetMathAlphabet{\mathsfit}{bold}{\encodingdefault}{\sfdefault}{bx}{n}

\newcommand{\E}{\mathbb{E}}

\DeclarePairedDelimiter\br{(}{)}
\DeclarePairedDelimiter\brs{[}{]}
\DeclarePairedDelimiter\abs{\lvert}{\rvert}

\DeclareMathOperator{\St}{\mathcal{S}}
\DeclareMathOperator{\A}{\mathcal{A}}

\usepackage[english]{babel}
\usepackage{amsthm}

\theoremstyle{plain}

\theoremstyle{definition}

\theoremstyle{remark}

\newtheorem*{theorem*}{Theorem} 
\newtheorem*{corollary*}{Corollary}
\newtheorem*{lemma*}{Lemma}
\newtheorem*{proposition*}{Proposition}

\usepackage{dsfont}

\usepackage{wrapfig}
\usepackage{sidecap}
\usepackage{float}
\usepackage{ulem}

\usepackage[accsupp]{axessibility}  

\def\paperID{11668}
\def\confName{CVPR}
\def\confYear{2025}

\title{RL-RC-DoT: A Block-level RL agent for Task-Aware Video Compression}
\newcommand{\ourmethod}{RL-RC-DoT}

\usepackage[symbol]{footmisc}  

\author{Uri Gadot\textsuperscript{1,2}
\and
Assaf Shocher \textsuperscript{2}
\and
Shie Mannor \textsuperscript{1,2}
\and
Gal Chechik \textsuperscript{2}
\and
Assaf Hallak \textsuperscript{2} \and
\textsuperscript{1}Technion \quad \textsuperscript{2}NVIDIA Research
\and \small Corresponding authors: \texttt{uri.gad@campus.technion.ac.il, ahallak@nvidia.com}
}

\begin{document}
\maketitle
\begin{abstract}
    Video encoders optimize compression for human perception by minimizing reconstruction error under bit-rate constraints. In many modern applications such as autonomous driving, an overwhelming majority of videos serve as input for AI systems performing tasks like object recognition or segmentation, rather than being watched by humans. It is therefore useful to optimize the encoder for a downstream task instead of for perceptual image quality. However, a major challenge is how to combine such downstream optimization with existing standard video encoders, which are highly efficient and popular. Here, we address this challenge by controlling the Quantization Parameters (QPs) at the macro-block level to optimize the downstream task. This granular control allows us to prioritize encoding for task-relevant regions within each frame. We formulate this optimization problem as a Reinforcement Learning (RL) task, where the agent learns to balance long-term implications of choosing QPs on both task performance and bit-rate constraints. Notably, our policy does not require the downstream task as an input during inference, making it suitable for streaming applications and edge devices such as vehicles. We demonstrate significant improvements in two tasks, car detection, and ROI (saliency) encoding. Our approach improves task performance for a given bit rate compared to traditional task agnostic encoding methods, paving the way for more efficient task-aware video compression.
\end{abstract}  
\section{Introduction}
\label{sec:intro}

Video compression is an essential and widely studied problem \citep{bhaskaran1997image, wenger2003h,sullivan2012overview,bross2021overview,kufa2017software}. Most video compression algorithms are designed for preserving how a video is perceived by people. With the success of computer vision applications, many videos are used in automated systems, from autonomous drones and cars, to security cameras, and in downstream tasks, like object detection or recognition. In these cases, compression must prioritize regions relevant to the task at hand (e.g., allocating more bits to objects than to the background). 

Real-world deployment of compression and computer vision systems complicates matters further. Video data must be collected in real time from devices, using low computational resources, and be usable for training various models across multiple tasks, not just for immediate inference. Furthermore, due to computational and hardware constraints, compression must be done without access to the ground truth for the downstream tasks during the encoding process.
Our goal is to tackle these challenges by providing a general video compression method that can be adapted to any task, any encoding standard, operates in real-time, imposes low computational demands on the encoding side, and requires no ground-truth labels.

Two previous approaches were taken to this problem. The first relies on standardized video encoders \citep{merritt2006x264, sullivan2012overview} which are highly efficient but are not designed for adapting compression to specific tasks in real-time. To address that, previous studies used downstream task information as input to the encoder side~\cite{li2021task, xie2022hierarchical, shi2020reinforced}. As one example, \cite{xie2022hierarchical} performs semantic compression by applying a heavy feature extractor before encoding using a ground-truth segmentation maps. This approach may compress  well at this setup, but typically require large computation resources before encoding, can not be used for various tasks, and is  unsuitable for data collection. A second approach relies on deep encoding \citep{lu2019dvc}. Once again it is computationally expensive and currently unsuitable for real-time applications or resource-constrained environments. 


In this paper, we propose \ourmethod{}, a novel solution to the problem of tuning an efficient real-time video compression system to a downstream task without access to its ground truth labels during inference. Our approach integrates a lightweight network on the video encoder side, trained to control the encoding process such that the decoded output is ideal for the task at hand. By leveraging standardized codecs, we ensure that our method is both computationally efficient and easily deployable across a range of devices. The solution allows for real-time video compression without requiring ground truth for downstream tasks. Our solution can be applied over any existing encoder, for simplicity, we chose to implement it over x264 as a mere example.

Coping with these challenges is hard. Standardized encoders are not differentiable, making it difficult to optimize bit allocation for specific tasks. To overcome this, we introduce a reinforcement learning (RL) mechanism that controls the Quantization Parameter (QP) at the macro-block (MB) level, adjusting the bit allocation for each block of the frame dynamically. This allows us to efficiently manage the bit-rate budget while optimizing task performance over an entire sequence of video frames. Our experiments demonstrate that this approach yields significant improvements in rate-distortion trade-offs, not just for the task the encoder was trained on, but also for other related tasks, showcasing the robustness of our method. Furthermore, we demonstrate its generalizability by showing how an encoder trained on one model can improve performance for other models without additional tuning.

In summary, this paper makes the following contributions. (1) We design the first task-aware video compression method that builds on top of existing encoders and does not require solving the task during inference. (2) We show how to optimize the quantization parameter of every MB in the frame while optimizing the performance of a downstream task on the reconstructed video under bit-rate constraints. (3) We design an architecture that outputs multiple actions, a tailored reward for this problem, and 
a task-prediction loss term. (4) We show improved rate-distortion trade-off for our agent on two tasks, car detection and ROI encoding with only small interference to image quality, and further show robustness to task shift, when tested on a related-but-different task than used for training.

\subsection{Related Works}
\textbf{Video compression with RL.}
The integration of reinforcement learning into video compression has gained significant attention in recent years. Several approaches have focused on frame-level QP optimization \cite{mandhane2022muzero, mao2020neural, ho2021dual, chen2018reinforcement}, using various techniques such as two-pass encoding and dual-critic architectures. While some researchers have proposed more granular control by developing QP selection agents at the Coding Tree Unit (CTU) or MacroBlock (MB) level \cite{ho2022neural, hu2018reinforcement}, these methods have been primarily validated only on image compression or intra-mode encoding scenarios.

\textbf{Task-aware video compression with unrestricted compute.}
Several previous studies proposed video compression methods that are aware of a downstream task. 
\cite{zhang2024competitive} explored 
content-specific filters to improve post-processing in video codecs, optimizing them for machine vision tasks like object detection and segmentation. \cite{ge2024task} introduced an encoder control for deep video compression that adapts to multiple tasks using a single pre-trained decoder, showing significant bit-rate improvement for object detection and tracking. \cite{shor2022need} highlighted the limitations of classical codecs in medical videos, proposing learned compression models to allocate more bits to medically relevant regions. \cite{elgamal2020sieve} presented 
a semantic video encoding system that enhances object detection by selectively 
decompressing frames in surveillance streams. \cite{li2024task} developed a distributed compression framework that adjusts to varying bandwidth in multi-sensor networks to optimize task performance. \cite{windsheimer2024annotation} introduced an annotation-free optimization strategy that aligns video coding with machine tasks, improving rate savings without relying on ground truth data. Additionally, While \cite{wu2024qs} focused on real-time, quality-scalable video decoding, it also evaluated the codec performance on machine-based tasks. 

All these approaches share a common limitation, they do not use the existing highly optimized and widely prevalent existing video compression ecosystem like the open-source x264 \citep{merritt2006x264}. The challenge therefore remains to design video compression systems that build on top of existing technology, but can be tuned in a-content adaptive way to a set of downstream tasks. 

\textbf{Task-aware video compression with standard encoders}
Another body of works does employ standardized encoders, but does not consider the inter-frame dependencies. \cite{singh2022video} and \cite{fischer2020video} optimize the CTU partitioning to improve the compression for a downstream task. \cite{galteri2018video} uses a threshold on the saliency map to allocate more bits to important regions, while \cite{cai2021novel} optimizes over the modelled relation between each block parameter and the task performance. \cite{li2021task} uses RL for optimizing block QPs, but does so in each frame separately, where the sequence is defined over the sequence of blocks in the same frame. In our work we output all MB QPs with one policy and the sequence is defined over consecutive encoded frames in the video. The work most related to ours is \cite{xie2022hierarchical}, where they propose to use RL on both the QPs and block QPs in a hierarchical manner. However, they limit their optimization to only two frames in every GOP, and only two values of block QPs are chosen per block according to a given segmentation map. In our work we optimize over all frames and MB QPs, and we do not use any additional information like saliency, segmentation or downstream task during inference.
\section{Preliminaries}
\label{sec:preliminaries}
\subsection{Video Compression}
\label{subsec:pre_video_compression}

Video compression is a process of reducing the size of digital video files while maintaining acceptable visual quality. It is a crucial technology in the modern digital age, enabling efficient storage, transmission, and streaming of video content across various platforms and devices. The primary goal of video compression is to eliminate redundant and less perceptible information from the video data according to constraints such as bit-rate of the target video.

One key aspect of video compression is the use of Quantization Parameters (QP). QP values control the level of compression and distortion applied to the video data, with higher values resulting in more compression but lower quality, and lower values preserving more detail but producing larger file sizes. In video encoding, QP can be applied at different levels of granularity. Frame QP refers to setting a single QP value for an entire frame, which is useful for maintaining consistent quality across the frame but may not be optimal for all areas. Per-block QP, conversely, allows for finer control by assigning different QP values to individual blocks within a frame, usually in small perturbations from a pre-assigned frame QP. This approach enables the encoder to apply more compression to less important or visually complex areas while preserving quality in critical regions. Per-MB QP can lead to more efficient compression and better overall visual quality, as it adapts to the local characteristics of the video content. It is especially suitable for task-aware optimization since most tasks target specific areas in the picture (e.g. object detection and segmentation).



One may wonder, if a downstream task is given, why is video compression needed at all? For instance, in the autonomous vehicle example, if a car detector is available, why not run that detector on the vehicle, and save only its decision instead of the compressed video. There are several strong reasons not to take this approach: (1) Many downstream tasks require resource-heavy networks that cannot run efficiently on-device, making it impractical to process the data locally. (2) Sending only task-specific features limits human interpretability, as there would be no watchable video for explainability. (3) This also confines the data to a single task, preventing its reuse for other applications or analyses. (4) Large-scale data collection, such as in autonomous driving, depends on compressed video storage; using features alone would limit future training and fine-tuning opportunities. (5) Task-specific features are often tied to a particular model, making them incompatible with new models, while compressed video remains adaptable across different systems. We show that our method allows different models to achieve high performance using the same compressed data. This is also the reason why we aim to develop a method that still preserves a video that would be meaningful to a person. 

\subsection{Reinforcement Learning}
Reinforcement Learning (RL; \cite{sutton1998introduction}) is a field dealing with sequential decision making in unknown environments. To formulate a problem using RL, we first need to define its underlying Markov Decision Process (MDP). An MDP is defined by a tuple $\br*{\St, \A, P, R, \gamma}$, where $\St$ is a finite set of states, $\A$ is a finite set of actions, $P$ is a state transition probability function, $P\br*{s'|s,a}$, $R$ is a reward function, $R\br*{s,a}$ and $\gamma \in \brs*{0,1}$ is a discount factor.\\
At each time step $t$, the agent observes the current state $s_t \in \St$ and chooses an action $a_t \in \A$. The environment then transitions to a new state $s_{t+1}$ with probability $P\br*{s_{t+1}|s_t,a_t}$ and the agent receives a reward $r_t = R\br*{s_t,a_t}$. The goal of the agent is to find a policy $\pi : \St \rightarrow \A$ that maximizes the expected cumulative discounted reward:
\begin{align*}
\max_{\pi} J^{\pi} = \E_{\pi,s_0\sim\mu, s_{t+1}\sim P}\brs*{\sum_{t=0}^{\infty}\gamma^tR\br*{s_t,\pi\br*{s_t}}}
\end{align*}

To do so, many algorithms were proposed in the literature varying in their assumptions on the problem, computational complexity and data requirements. Perhaps the most widely used algorithm today is PPO \citep{schulman2017proximal} which directly optimizes the policy using full trajectories while constraining it from diverging. 

When the action space is high-dimensional, as in our case, learning becomes exponentially harder. This can lead to extremely slow learning progress, and requires function approximators \cite{van2007reinforcement}, which are prone to overfitting and generalize poorly. Consequently, techniques specifically designed for high-dimensional action spaces, such as action abstraction, or dimensionality reduction \cite{pierrot2021factored}, are often necessary to make the problem tractable and enable efficient reinforcement learning.

\section{Method} \label{sec:method}

\begin{figure*}[t]
\centering
\includegraphics[width=0.9\linewidth]{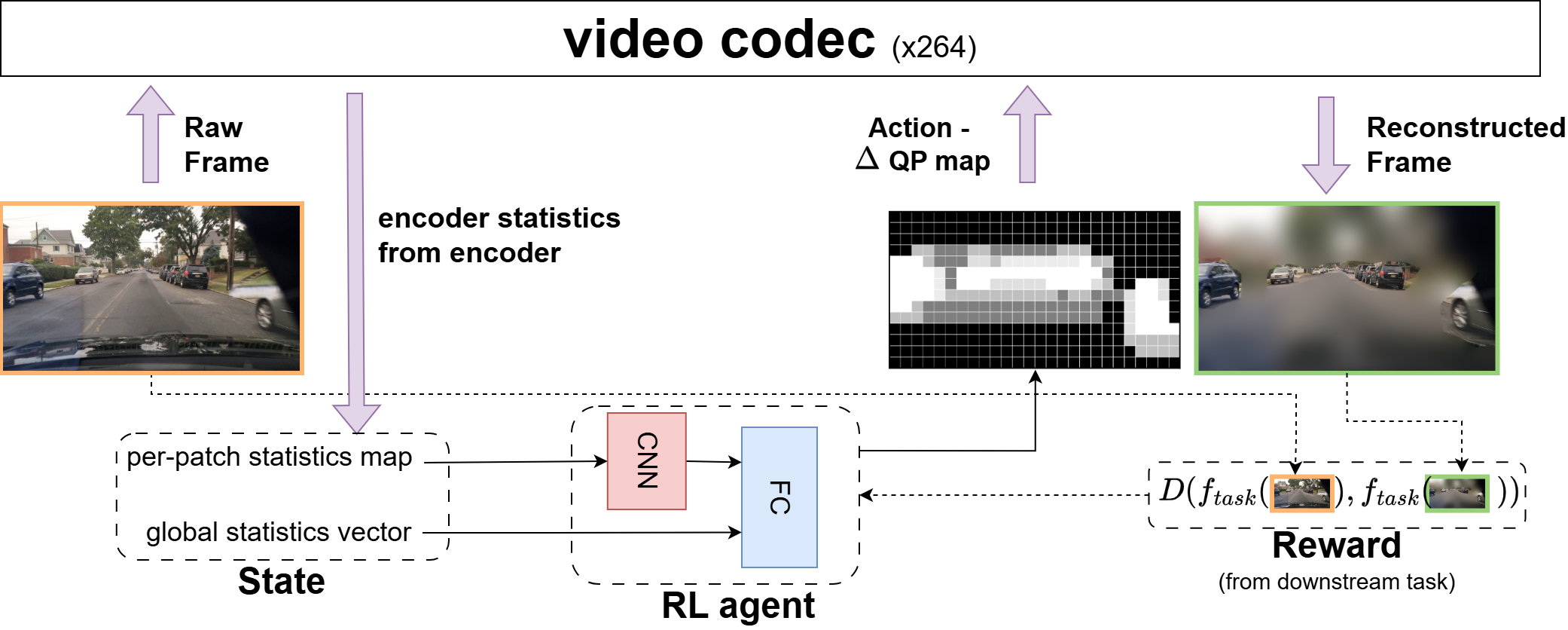}  
    \caption{\textbf{\ourmethod{} workflow.} Our proposed solution to the block-level control for a downstream task. \ourmethod{} takes encoder statistics as input and outputs a block-level delta QP map. We then evaluate the difference in downstream task performance between the reconstructed frame and the raw frame. The reward contains both global score as reward and block-level score.}
\label{fig:method}
\end{figure*}

We present \ourmethod{}, an \textbf{RL}-based \textbf{R}ate \textbf{C}ontroller for \textbf{Do}wnstream \textbf{T}ask, that dynamically optimizes pet-block QP deltas during video encoding. Simply put, our approach is to train a deep-RL agent that allocate bandwidth to different patches in each frame, only observing fast-to-compute features of every frame. To formalize the training framework, we cast the video compression problem with respect to a downstream task as a Markov decision process (MDP). In our formulation, each time-step of the MDP corresponds to processing one frame of the video. To train the policy, we use the PPO algorithm \citep{schulman2017proximal}.

\subsection{The MDP model}
We now describe the various components of our model. 

\textbf{State space:} We define the state of the environment to be per-patch statistics extracted from the encoder. The specific statistics we used are described below and in Appendix~\ref{sec:apx_env_details}. 

%

\noindent \textbf{Action space:} We define an action to be a choice of all MB QP deltas within a frame. For example, given a frame resolution of 480×320 pixels partitioned into 16×16 pixel blocks, the resulting action space constitutes a 30×20 dimensional matrix, where each value represent the delta between the frame level QP and the block specific QP. This action space is very high-dimensional and we discuss below  an efficiency improvement.

\noindent 
\textbf{Reward: }We define the reward as a combination of two rewards with different purposes. 

First, we wish to maximize the performance of the downstream task on decoded frames. Since ground-truth data is unavailable, we introduce a novel self-supervised approach. This method treats the downstream task's output on the original uncompressed frame as a pseudo-ground-truth, against which we evaluate the task performance on the reconstructed frame:
\begin{equation*}
    r_\text{task} = D\left[f_\text{task}(\text{frame}_\text{raw}), f_\text{task}(\text{frame}_\text{recon})\right],
\end{equation*}
where $f_\text{task}$ is a pre-trained model for the downstream-task and $D$ is a task-specific objective function. For example, for car detection, $f_{task}$ we use a pre-trained car detection model (YOLO-v5 nano \citep{yolov5}), and $D$ measures how well the downstream task model of $f_{task}(\text{frame}_\text{recon})$ performs compared to the raw data $f_{task}(\text{frame}_\text{raw})$.
Here, $f_{task}(\text{frame}_\text{raw})$ is used as a pseudo-ground-truth, because we may not have access to the ground-truth of the frame.

The second component of the reward is designed for complying with the encoder's bit-rate constraint. This is particularly crucial in streaming applications, where exceeding the allocated bandwidth can result in frame dropping and consequently deteriorate the viewer experience. The reward component $r_{\text{bit-rate}}$ for this objective is:
\begin{equation*}
    r_\text{bit-rate} = - \abs*{\log{\br*{\frac{\text{current average bit-rate}}{\text{target bit-rate}}}}}.
\end{equation*}
The final reward is a weighted combination of the two components: 
$$r = r_\text{bit-rate} + \lambda r_\text{task}, $$ where $\lambda$ is a hyper-parameter that determines rate-performance trade-off.

\paragraph{Efficient action space:}
The naive action space described above is very high dimensional, presenting a significant computational challenge in term of convergence of reinforcement learning algorithms. To address this difficulty, we implement a coarsening approach: the agent operates on a lower-resolution action space, which is subsequently up-sampled to the original dimensions through interpolation. This technique facilitates more efficient training while maintaining the ability to generate fine-grained QP assignments. We analyze the impact of action space resolution on model performance in Appendix~\ref{subsec:mbmult}.

Finally, unlike previous approaches ~\cite{li2021task, xie2022hierarchical} that relied on complex downstream task information (e.g. semantic maps, grad-cam), our method simplifies the state space by utilizing only encoder statistics that are readily available in the x264 encoder, so no additional calculations are required. A diagram of the full system is given in Figure~\ref{fig:method} and additional details on the architecture used by the PPO algorithm is given in Appendix ~\ref{appendix:architecture}.

\begin{SCtable*}[][t]
    \centering
    \setlength{\tabcolsep}{3pt}
    \begin{tabular}{@{}lcccccc@{}}
    \toprule
      & \multicolumn{2}{c}{\textbf{Precision}} & \multicolumn{2}{c}{\textbf{Recall}} & \multicolumn{2}{c}{\textbf{PSNR}}\\
      \textbf{}  & \textbf{low-rate} &  \textbf{high-rate}& \textbf{low-rate} &  \textbf{high-rate}& \textbf{low-rate} &  \textbf{high-rate} \\
    \midrule
     x264 & $.22\pm .0013$ & $.66 \pm .0015$ & $.45 \pm .002$ & $.81 \pm .001$ & $28.98\pm .03$ &  $34.55 \pm .03$ \\
    \ourmethod{} & \textbf{.36$\pm$ .0015} & \textbf{.71 $\pm$ .0014} & \textbf{.63 $\pm$ .002} & \textbf{.83$\pm$ .001} & {$29.03\pm .03$}&  {$34.55 \pm .03$} \\
    \bottomrule
    \end{tabular}
    \caption{Car detection precision and recall of YOLO5, and PSNR. Value are mean and s.e.m. calculated across all frames from the test set.}
    \label{tab:per_rate_mean_results}
\end{SCtable*}
\subsection{Macro-Block Reward Information}\label{subsec:MBRI}
In most RL problems, the reward is a black-box directly mapping the state to a continuous score. Recent literature~\citep{ye2021mastering} has demonstrated that predictive modeling of rewards can significantly enhance agent performance. more specifically they implemented it as auxiliary heads alongside policy or value networks.
In our setup, the reward presents a unique characteristic: the reward signals for various downstream tasks are often compositional. This occurs because the scores are derived from aggregating measurements across local patches in each input frame. For example, when optimizing for saliency-weighted PSNR, the reward is computed by aggregating per-pixel reconstruction errors. 

To leverage this decomposable nature of rewards, we propose to add 
an auxiliary prediction loss for these sub-scores during back-propagation. Specifically, we introduce a block-wise loss that aims to predict for each individual block, the local reward that contribute to the overall task score. Adding this loss for macro-block level reward information is expected to enhance the agent's performance. First, it provides a more localized learning signal, allowing the agent to understand the impact of its actions on individual components of the reward. Second, by learning to predict these sub-scores, the agent develops a richer internal representation of the task structure. Lastly, this method aligns the agent's learning more closely with the actual composition of the reward, potentially leading to faster convergence and more stable learning. Section \ref{sec:ablation} shows the effect of this improvement.


\section{Experiments}
\label{sec:exps}


We evaluate our approach with two downstream tasks: car detection and region-of-interest (ROI) encoding \citep{liu2008region}. We further study the robustness of the method, when a trained compression policy is tested with a different car detector, or even in a segmentation task instead of detection. Finally, we report performance of ablation experiments.



\subsection{Dataset}\label{subsec:dataset}
We trained and evaluated \ourmethod{} using a subset of video streams from the BDD100K dataset \citep{yu2020bdd100k}, a large-scale driving video dataset, with multi-task annotations. We reconstructed the raw data from the videos and to allow faster training time, we resized them to a smaller resolution of 480x320 pixels. We then filtered out streams that exhibited trivial rate-task performance (RD) curves with respect to the downstream tasks of car detection precision, when encoded with the standard x264 codec \citep{wiegand2003overview}. We specifically excluded streams that showed zero precision across most target bit-rates. This approach ensured that our dataset presented meaningful challenges for compression optimization. 

Our final dataset comprised of 172 streams in total, with 65 streams used for training our agent, 7 streams used for evaluation on different hyper-parameters and 100 streams reserved for testing. For reproducibility, we provided a detailed list of the specific stream used in our experiments in Appendix~\ref{sec:bdd100k} of this paper.

\subsection{Evaluation metrics: RD-curve and BD-rate}
\label{subsec:rd_bd}
\begin{SCfigure*}[][t]
    \centering
    \includegraphics[width=0.75\textwidth]{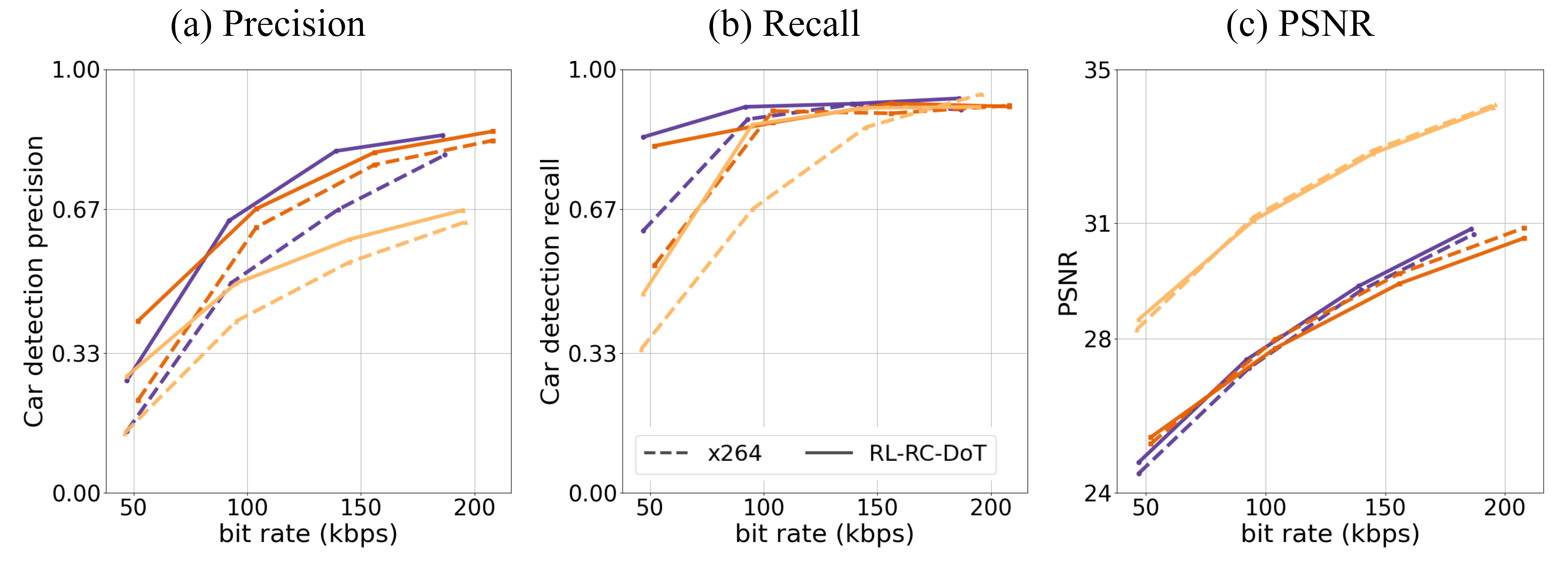}
     \caption{RD curves for Car detection task for different streams (color). Comparing standard x264 (dashed lines)  with \ourmethod{} (solid lines). Curves show 3 example streams, demonstrating how \ourmethod{} improves quality across the range of bit-rate values.  \textbf{(a)} Car detection precision \textbf{(b)} recall \textbf{(c)} PSNR. }
    \label{fig:rd_curves}
\end{SCfigure*}

Since compression is a constraint optimization problem, it is standard to depict results using a Rate-Distortion (RD) curve. Single RD-curve illustrates the trade-off between bit-rate constraint and quality in video compression (see Figure \ref{fig:rd_curves}). RD-curves are traditionally used with PSNR, but are equally applicable to task-specific metrics like precision/recall for a detection task or saliency-weighted PSNR for ROI-based encoding. These RD-curves allow us to evaluate  compression efficiency for any downstream tasks on reconstructed videos. 

BD-rate (Bjøntegaard Delta rate; \citep{Bjntegaard2001CalculationOA}) is a widely used metric in video compression to compare the efficiency of different encoding methods. This method calculates the average difference in bit-rate between two rate-distortion (RD) curves at the same quality level. The BD-rate represents the percentage of bit-rate savings that one encoding method achieves over another while maintaining equivalent video quality  performance. Thus, a negative BD-rate indicates that the test method requires less bits than the reference method to achieve the same quality / task performance.

\subsection{Compared methods: }To conduct a fair and meaningful comparison against existing baselines, they should be solving the same task, and particularly have access to the same information. Several previous studies developed methods for task-aware encoding, but their setup is fundamentally different. For instance, some previous methods assume access to the downstream model during inference \cite{xie2022hierarchical} resulting in a much higher computational cost depending on the task.
\cite{li2021task} and \cite{fischer2020video} focus on a single-frame (image) compression, ignoring the overall video budget constraints. Finally, most methods did not release code \cite{shi2020reinforced,fischer2020video}. These differences in approach and constraints make direct comparisons misleading.  


\subsection{Experimental details}\label{subsec:exp_details}
All of our experiments use the x264 open source encoder software \citep{merritt2006x264}, with the \textit{medium} preset and target bit-rates $50-200$ kbps. To extract the statistics we allow x264 to use look-ahead for $10$ frames. For car detection, we employ YOLOv5-nano \citep{yolov5}. ROI encoding is evaluated using saliency maps generated by TranSalNet \citep{TranSalNet}. Our agent is trained using Stable-Baselines3 \citep{raffin2021stable} implemented PPO  with the reward function described in Section \ref{sec:method}. We augment the standard PPO algorithm with a reward per block prediction network, as described in Section \ref{subsec:MBRI}. 
The computing time of \ourmethod{} is 0.004 sec, equivalent to 250FPS. This is added on top of the compute time of the encoder. It ensures that our agent is suitable for real-time video compression.


\section{Results}
\subsection{Car Detection}
\label{sec:car_detection}

We first assess the performance of \ourmethod{} in the context of video compression optimized for car detection. The reward function for training our RL agent is based on the precision score of YOLOv5-nano~\citep{yolov5}. In practice it was calculated as true positives divided by total positives in the granularity of the detection boxes. For our additional auxiliary loss described in Section \ref{subsec:MBRI}, we compute the precision score for each individual block separately to generate block-specific reward information.

Table \ref{tab:per_rate_mean_results} compares  \ourmethod{} with the standard x264 encoder, focusing on the detection performance of the YOLOv5-nano detector on compressed videos. The evaluation is conducted across multiple compression rates, with results averaged over all frames in the test dataset for each target bit-rate. We also applied the same comparative approach to assess the PSNR of the reconstructed streams. The results demonstrating that \ourmethod{} improves car detection precision and recall significantly, with minimal impact on the PSNR of the compressed videos. 

\begin{SCfigure*}[][t]
     \centering
     \includegraphics[width=0.7\textwidth]{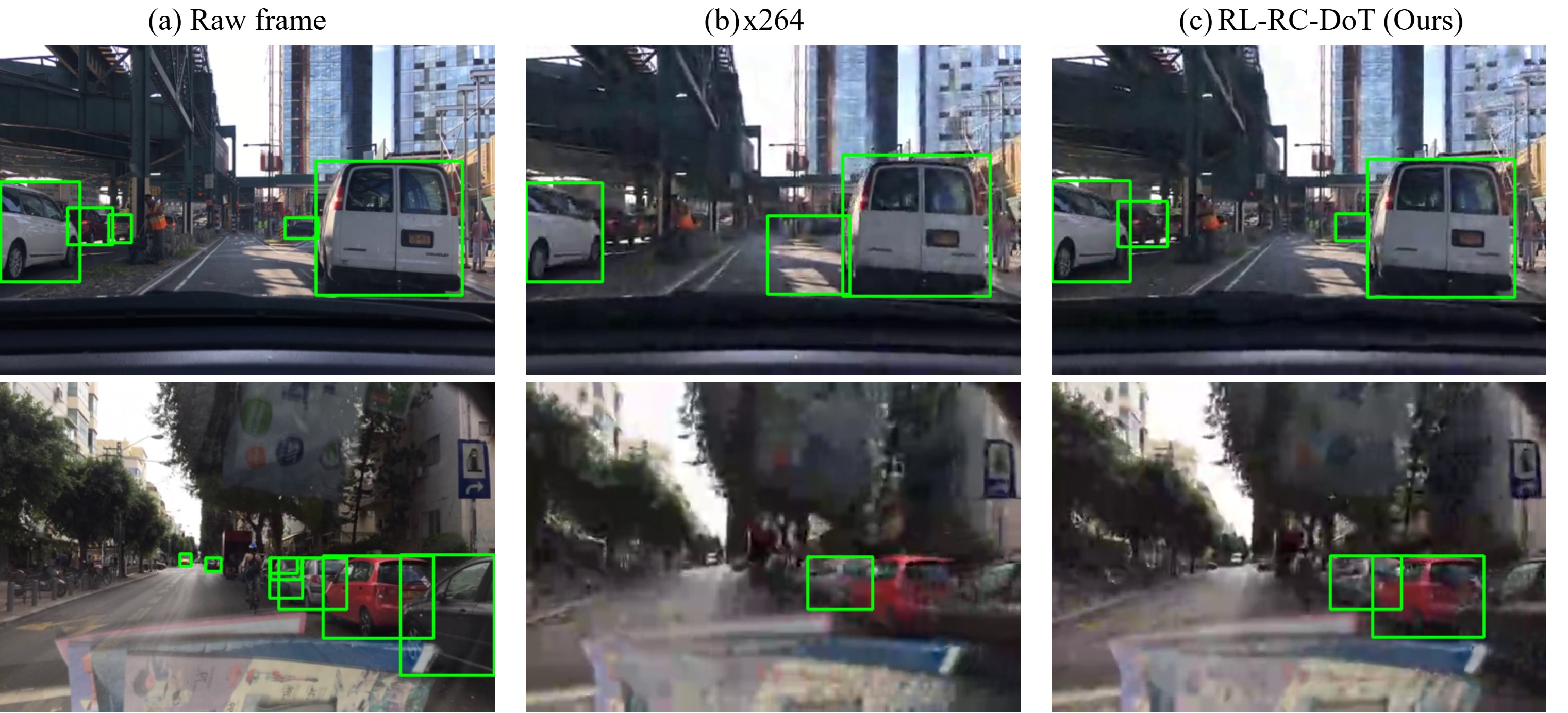}
        \caption{Car detection example result. (a) detection output on x264 reconstructed frame, (b) output on raw frame and (c) output on \ourmethod{} reconstructed frame. Notice that both \ourmethod{} and x264 used the same target bit-rate}
        \label{fig:car_detection_frames}
\end{SCfigure*}

Figure \ref{fig:rd_curves} illustrates the superiority of \ourmethod{} over the standard x264 encoder through RD curves for three representative video streams. Figure \ref{fig:car_detection_frames}, shows a qualitative example of the performance gain. We compare the images in both types of rate-control, and the output of the downstream task. We can see the details corresponding to the downstream task are better reconstructed yielding a more relevant image.  

To quantify the performance difference between methods, we compute the BD-rate (see \ref{subsec:rd_bd}), a standard metric in the field. 
Our approach shows significant improvements in detection performance, with BD-rate reductions of $24.7\% (\pm 1.38\%)$. These gains come at a minimal cost to overall video quality, yielding a slight increase (deterioration) in PSNR BD-rate of $1.19\% (\pm 0.46\%)$ (Table \ref{tab:car-detection}). This means that videos compressed using \ourmethod{} remains understandable to human viewers, a crucial aspect for validation and debug purposes. It also exhibits robustness to different task models, which we elaborate on in section \ref{subsec:task_robustness}. 

\subsection{Quantization map analysis}
\begin{SCfigure*}[][t]
    \centering
    \includegraphics[width=0.7\textwidth]{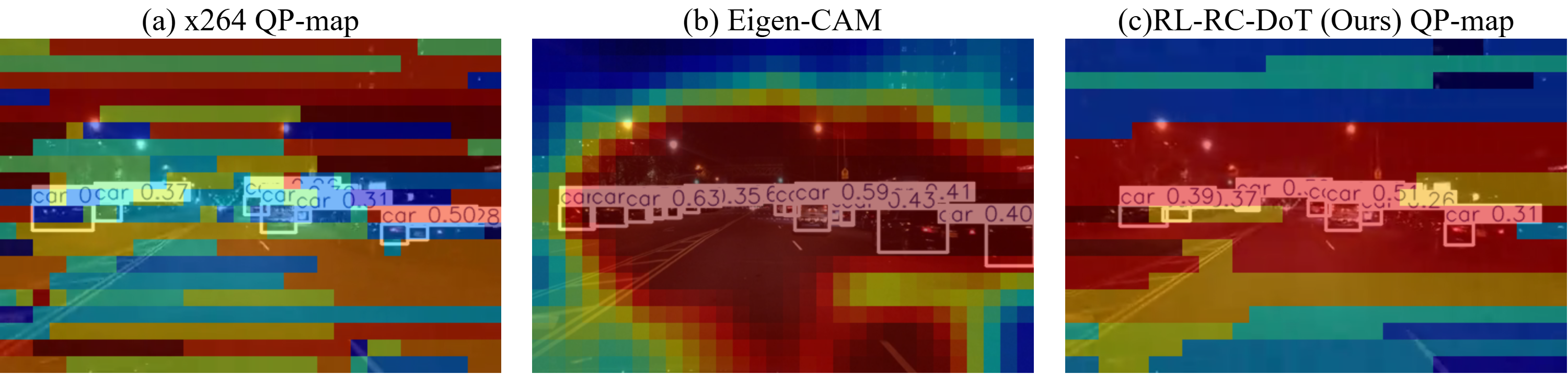}
     \caption{Example frame analysis. Red areas denote more bits allocate in the QP map (\textbf{(a)} x264 allocated bits,\textbf{(c)} \ourmethod{} allocated bits ) and \textbf{(b)} Eigen-CAM values (middle).}
    \label{fig:qp_maps}
\end{SCfigure*}

To gain more insight into how our approach affects QP map, we quantified the relation between the QP map and areas in the video that are useful for car detection. Specifically, we computed the KL-divergence between normalized QP-maps and Eigen-CAM \cite{muhammad2020eigen} spatial map. Figure \ref{fig:qp_maps} illustrates the three maps for one frame, showing that RL-RC-DoT allocates more bits to areas in the frame that are informative for car detection. Figure~\ref{fig:qpmaps-stat} (provided in the supplementary material) aggregates this metric across our entire test dataset, confirming our approach's effectiveness with significantly lower KL-divergence values (mean $D_{kl}$ for RL-RC-DoT = $2.6$ compared to x264=$4.4$).

\subsection{ROI encoding}

\begin{figure}[htp]
     \centering
     \includegraphics[width=0.65\linewidth]{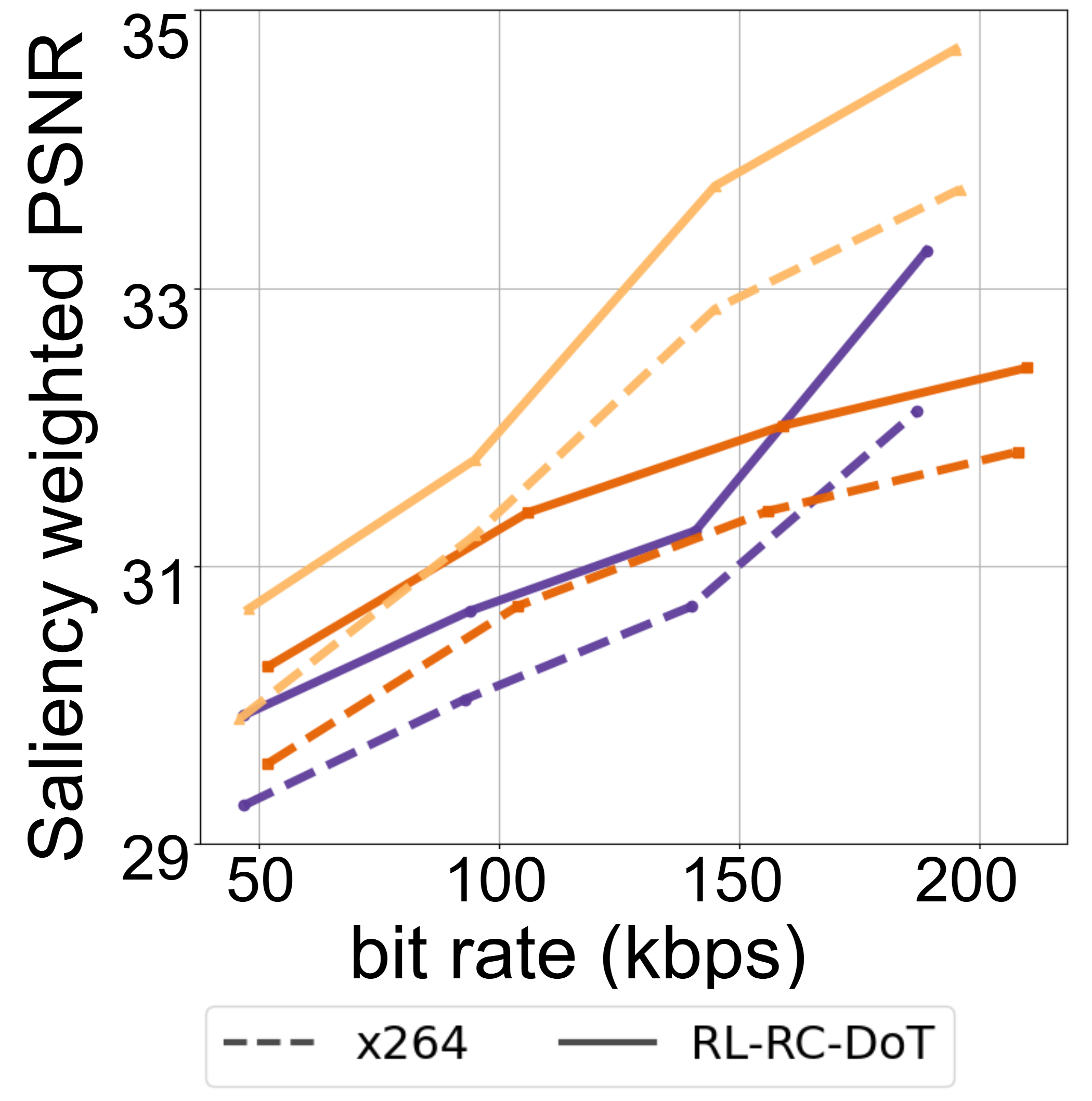}     
    \caption{RD-curves for 3 videos (color) for ROI-encoding. }
    \label{fig:roi_rd_curves}
\end{figure}

\begin{SCfigure*}[][t]
     \centering
     \includegraphics[width=0.7\textwidth]{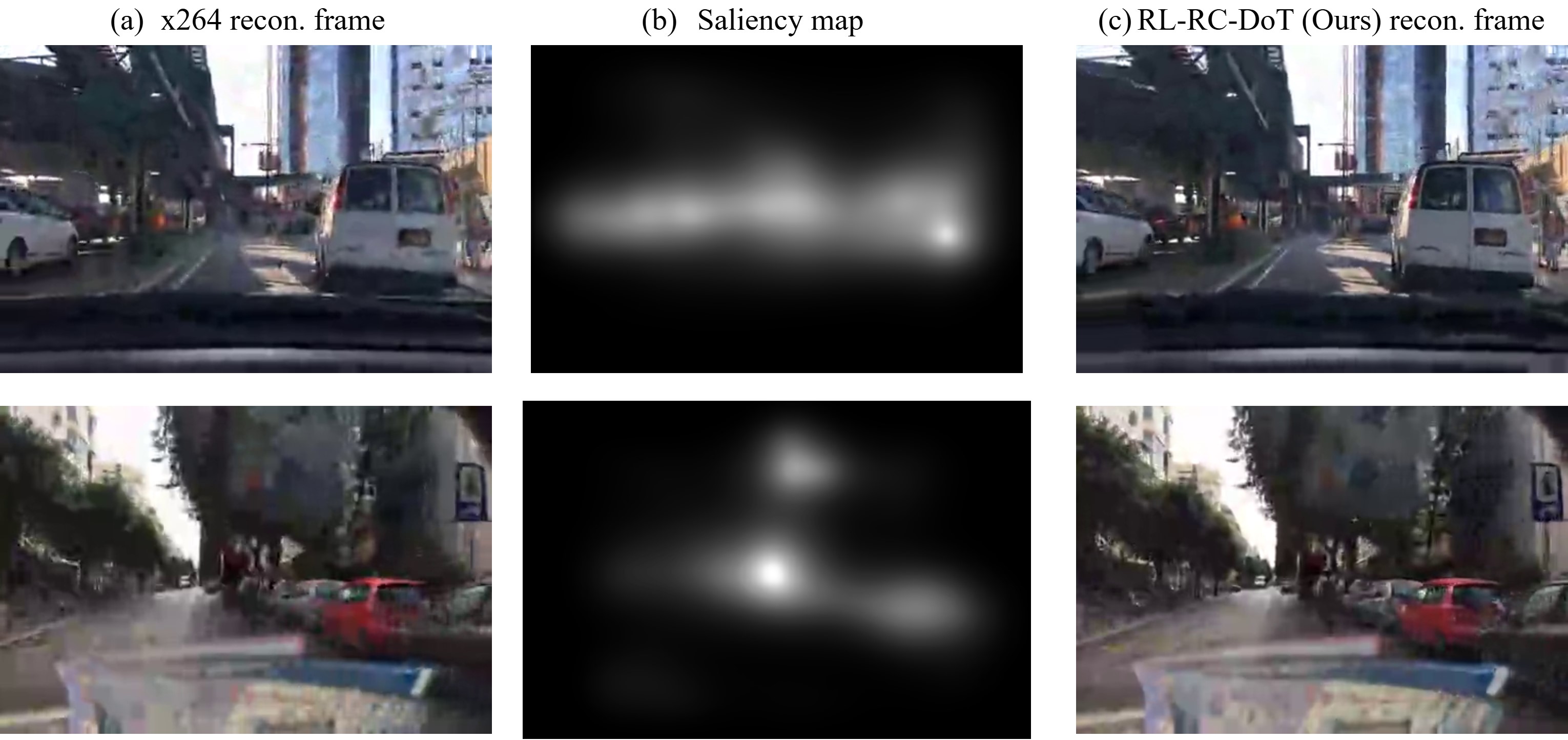}
        \caption{Saliency weighted PSNR results. (a) x264 reconstructed frame, (b) Saliency map of raw frame, extracted with ~\citep{TranSalNet} (c) \ourmethod{} reconstructed frame. Notice that both \ourmethod{} and x264 used the same target bit-rate}
        \label{fig:car_saliency_frames}
\end{SCfigure*}

We conducted similar experiments for ROI-encoding task by promoting saliency weighted PSNR as the task score.
\ourmethod{} exhibits a BD-rate value of $-25.64\% (\pm0.99\%)$, indicating that our method achieves significantly better quality in salient regions at lower bit-rates compared to x264. Interestingly, the PSNR BD-rate obtained is $-5.26 \pm 0.36$ which is slightly better than the vanilla rate-control. This may be due to the proximity between the two tasks. This also shows the sub-optimality of the vanilla rate-control when considering specific content. 
Figure \ref{fig:roi_rd_curves} illustrates the RD curves for three representative video streams. These curves demonstrate that in most cases, \ourmethod{} achieves a more favorable RD trade-off for ROI encoding task compared to x264. Figure \ref{fig:car_saliency_frames} provides qualitative examples of our method's performance, visually illustrating the enhanced quality in salient regions compared to the baseline encoding.

\subsection{Task Robustness}
\label{subsec:task_robustness}
An important concern is that \ourmethod{} might overfit for the training task. That would mean that changing the model, may harshly hurt performance. We set to evaluate robustness to such changes in \ourmethod{} by training the policy with one downstream task ,and testing it with another. 
More specifically, we optimized the policy for car detection using the YOLOv5-nano model, as described in section \ref{sec:car_detection}.
Then, we measured the detection performance  of another model, SSD \citep{liu2016ssd}. We also measure the performance on the related but distinct task of car segmentation (DeepLab; \citep{chen2017deeplab}). 
The results are also listed in Table \ref{tab:car-detection}.

\begin{table}[h]
    \centering
    \begin{tabular}{@{}ccc@{}}
    \toprule
     \textbf{Precision} & \textbf{Recall} &\textbf{PSNR}     \\
     \textbf{(YOLO)} & \textbf{(YOLO)}\\
    \midrule
    $-24.7 \pm 1.57$ & $-19.75 \pm 2.97$ & $1.19 \pm 0.46$\\ 
    \toprule
     \textbf{Precision}& \textbf{Recall} & \textbf{Segmentation} \\
       \textbf{(SSD)}& \textbf{(SSD)} &  \textbf{IOU} \\
         \midrule
     $-26.2 \pm 1.48$& $-25.81 \pm 2.03$ & $-14.6 \pm 1.81$\\ 
    \bottomrule
    \end{tabular}
        \caption{
        BD-rate Results on \ourmethod{} applied on test set for the car detection task for various settings. Negative values mean that \ourmethod{} improves over baseline.}
    \label{tab:car-detection}
\end{table}
 
This approach allows us to examine whether our method truly captures fundamental aspects of visual information relevant to automotive perception tasks, rather than overfitting to a specific model or narrow task definition. By demonstrating performance improvements across different models and related tasks, we aim to show that our compression method preserves task-relevant information in a more general sense, potentially allowing for model updates or task modifications without the need to retrain the compression policy. This robustness is crucial for real-world applications where deployed systems may need to adapt to new models or slightly different tasks over time. 
 
For car detection evaluated with SSD, the precision BD-rate is very similar to  precision with YOLOv5-nano, which was used for training. For car segmentation, although tested with a different task, we still observe an improved but weaker BD-rate than the detection task. This improvement can be attributed to the close relation between the tasks, so meaningful macro-blocks for car detection, are also useful for the segmentation task. In summary, the BD-rate obtained on the PSNR and the various tasks show the robustness of our method to new tasks and new models that solve the task. Additional qualitative experiments examining task robustness are detailed in the appendix

\begin{table}[htb!]
    \centering
    \begin{tabular}{lcc}
    \toprule 
    \textbf{} & \textbf{Precision} & \textbf{PSNR}\\
    \textbf{Car detection} & \textbf{BD-rate} & \textbf{BD-rate} \\
    \midrule
    \ourmethod{} & $-24.7 \pm 1.57$ & $1.19 \pm 0.46$ \\
    \ourmethod{} w/o RI & $-19.4 \pm 1.38$  & $1.92 \pm 0.41$  \\
    \ourmethod{} $\gamma=0$ & $-9.78 \pm 1.29$  & $5.44 \pm 0.6$ \\
    \midrule
    \textbf{} & \textbf{Sal.-weighted} & \textbf{PSNR} \\
    \textbf{ROI encoding} & \textbf{PSNR BD-rate} & \textbf{BD-rate}\\
    \midrule
    \ourmethod{} & $-25.64 \pm 0.99$ & $-5.26 \pm 0.36$ \\
    \ourmethod{} w/o RI & $-23.46 \pm 0.97$ & $-4.54 \pm 0.42$\\
    \ourmethod{} $\gamma=0$ & $-16.01 \pm 0.77$ & $2.11 \pm 0.31$\\
    \bottomrule
    \end{tabular}
    \caption{Ablation study. (1) Full \ourmethod{} (2) Omitting reward information (RI) from the training process and (3) Ignoring long term effects by using a myopic policy.} 
    \label{tab:ablation}
\end{table}

\subsection{Ablation Experiments} \label{sec:ablation}
To quantify the relative contribution of various components of our method, we perform ablation studies, for both car detection and ROI encoding, and provide the results in Table \ref{tab:ablation}. For both tasks, we first ablated the macro-block reward information as described in Subsection \ref{subsec:MBRI}. Then, ran an experiment for $\gamma=0$ which shows what happens when optimizing for a myopic policy. 

The results show that reward info improved the learning process and reduces the BD-rate even further for both tasks. This demonstrates the benefit of exploiting additional information in the video compression domain that is generally not available. For $\gamma=0$, the BD-rate is significantly worse for both tasks. As expected, ignoring the future implications of the bit-allocation can cause sub-optimal decisions for the entire video. This also emphasizes the limitation of rate-control methods optimizing for every frame separately; a common practice by previous works. 

\section{Conclusions and Limitations}
ML for video understanding is widespread but costly to store, making efficient compression essential. Current task-aware compression methods face limitations like computational demands and ground truth task data dependencies. We develop an efficient RL solution which encodes every frame in real time while optimizing the future bit-rate and task performance on the reconstructed video. Our learned policy is robust against changes in the downstream models for the same task and to closely related tasks, showing large important potential for data collection for autonomous vehicle, patient monitoring and robotics. 

In this work, we focused on H.264 due to its widespread use, but our mechanism is inherently codec-agnostic. Extending it to H.265 or H.266 would require adaptations, such as incorporating encoder-specific features like CTU structures.  Additionally, integrating unsupervised learning metrics into the reward function could improve task generalization by enabling the agent to learn more transferable representations of video quality. We propose these directions as promising avenues for future work.

\textbf{Limitations:}
Training our models involves encoding and performing the downstream task per frame, and this may slow down converge depending on the complexity of the downstream task. Also, generalizing across video resolutions may be hard because it affects the size of action space and the complexity of the learning problem.

\clearpage
{
    \small
    \bibliographystyle{ieeenat_fullname}
    \bibliography{main}

\begin{thebibliography}{48}
\providecommand{\natexlab}[1]{#1}
\providecommand{\url}[1]{\texttt{#1}}
\expandafter\ifx\csname urlstyle\endcsname\relax
  \providecommand{\doi}[1]{doi: #1}\else
  \providecommand{\doi}{doi: \begingroup \urlstyle{rm}\Url}\fi

\bibitem[Bhaskaran and Konstantinides(1997)]{bhaskaran1997image}
Vasudev Bhaskaran and Konstantinos Konstantinides.
\newblock Image and video compression standards: algorithms and architectures.
\newblock 1997.

\bibitem[Bj{\o}ntegaard(2001)]{Bjntegaard2001CalculationOA}
Gisle Bj{\o}ntegaard.
\newblock Calculation of average psnr differences between rd-curves.
\newblock 2001.

\bibitem[Bross et~al.(2021)Bross, Wang, Ye, Liu, Chen, Sullivan, and Ohm]{bross2021overview}
Benjamin Bross, Ye-Kui Wang, Yan Ye, Shan Liu, Jianle Chen, Gary~J Sullivan, and Jens-Rainer Ohm.
\newblock Overview of the versatile video coding (vvc) standard and its applications.
\newblock \emph{IEEE Transactions on Circuits and Systems for Video Technology}, 31\penalty0 (10):\penalty0 3736--3764, 2021.

\bibitem[Cai et~al.(2021)Cai, Chen, Wu, Liu, and Li]{cai2021novel}
Qi Cai, Zhifeng Chen, Dapeng~Oliver Wu, Shan Liu, and Xiang Li.
\newblock A novel video coding strategy in hevc for object detection.
\newblock \emph{IEEE Transactions on Circuits and Systems for Video Technology}, 31\penalty0 (12):\penalty0 4924--4937, 2021.

\bibitem[Chen et~al.(2017)Chen, Papandreou, Kokkinos, Murphy, and Yuille]{chen2017deeplab}
Liang-Chieh Chen, George Papandreou, Iasonas Kokkinos, Kevin Murphy, and Alan~L Yuille.
\newblock Deeplab: Semantic image segmentation with deep convolutional nets, atrous convolution, and fully connected crfs.
\newblock \emph{IEEE transactions on pattern analysis and machine intelligence}, 40\penalty0 (4):\penalty0 834--848, 2017.

\bibitem[Chen et~al.(2018)Chen, Hu, and Peng]{chen2018reinforcement}
Lian-Ching Chen, Jun-Hao Hu, and Wen-Hsiao Peng.
\newblock Reinforcement learning for hevc/h. 265 frame-level bit allocation.
\newblock In \emph{2018 IEEE 23rd International Conference on Digital Signal Processing (DSP)}, pages 1--5. IEEE, 2018.

\bibitem[Elgamal et~al.(2020)Elgamal, Shi, Gupta, Jana, and Nahrstedt]{elgamal2020sieve}
Tarek Elgamal, Shu Shi, Varun Gupta, Rittwik Jana, and Klara Nahrstedt.
\newblock Sieve: Semantically encoded video analytics on edge and cloud.
\newblock In \emph{2020 IEEE 40th International Conference on Distributed Computing Systems (ICDCS)}, pages 1383--1388. IEEE, 2020.

\bibitem[Fischer et~al.(2020)Fischer, Brand, Herglotz, and Kaup]{fischer2020video}
Kristian Fischer, Fabian Brand, Christian Herglotz, and Andr{\'e} Kaup.
\newblock Video coding for machines with feature-based rate-distortion optimization.
\newblock In \emph{2020 IEEE 22nd International Workshop on Multimedia Signal Processing (MMSP)}, pages 1--6. IEEE, 2020.

\bibitem[Galteri et~al.(2018)Galteri, Bertini, Seidenari, and Del~Bimbo]{galteri2018video}
Leonardo Galteri, Marco Bertini, Lorenzo Seidenari, and Alberto Del~Bimbo.
\newblock Video compression for object detection algorithms.
\newblock In \emph{2018 24th International Conference on Pattern Recognition (ICPR)}, pages 3007--3012. IEEE, 2018.

\bibitem[Ge et~al.(2024)Ge, Luo, Zhang, Xu, Lu, He, Geng, Wang, Zhang, and Qin]{ge2024task}
Xingtong Ge, Jixiang Luo, Xinjie Zhang, Tongda Xu, Guo Lu, Dailan He, Jing Geng, Yan Wang, Jun Zhang, and Hongwei Qin.
\newblock Task-aware encoder control for deep video compression.
\newblock In \emph{Proceedings of the IEEE/CVF Conference on Computer Vision and Pattern Recognition}, pages 26036--26045, 2024.

\bibitem[He et~al.(2017)He, Gkioxari, Doll{\'a}r, and Girshick]{he2017mask}
Kaiming He, Georgia Gkioxari, Piotr Doll{\'a}r, and Ross Girshick.
\newblock Mask r-cnn.
\newblock In \emph{Proceedings of the IEEE international conference on computer vision}, pages 2961--2969, 2017.

\bibitem[Ho et~al.(2021)Ho, Jin, Liang, Peng, and Li]{ho2021dual}
Yung-Han Ho, Guo-Lun Jin, Yun Liang, Wen-Hsiao Peng, and Xiaobo Li.
\newblock A dual-critic reinforcement learning framework for frame-level bit allocation in hevc/h. 265.
\newblock In \emph{2021 Data compression conference (DCC)}, pages 13--22. IEEE, 2021.

\bibitem[Ho et~al.(2022)Ho, Kao, Peng, and Hsieh]{ho2022neural}
Yung-Han Ho, Chia-Hao Kao, Wen-Hsiao Peng, and Ping-Chun Hsieh.
\newblock Neural frank-wolfe policy optimization for region-of-interest intra-frame coding with hevc/h. 265.
\newblock In \emph{2022 IEEE International Conference on Visual Communications and Image Processing (VCIP)}, pages 1--5. IEEE, 2022.

\bibitem[Hu et~al.(2018)Hu, Peng, and Chung]{hu2018reinforcement}
Jun-Hao Hu, Wen-Hsiao Peng, and Chia-Hua Chung.
\newblock Reinforcement learning for hevc/h. 265 intra-frame rate control.
\newblock In \emph{2018 IEEE International Symposium on Circuits and Systems (ISCAS)}, pages 1--5. IEEE, 2018.

\bibitem[Jocher(2020)]{yolov5}
Glenn Jocher.
\newblock {ultralytics/yolov5}.
\newblock \url{https://github.com/ultralytics/ultralytics}, 2020.

\bibitem[Kong et~al.(2016)Kong, Dai, and Zhang]{kong2016new}
Lingchao Kong, Rui Dai, and Yuchi Zhang.
\newblock A new quality model for object detection using compressed videos.
\newblock In \emph{2016 IEEE International Conference on Image Processing (ICIP)}, pages 3797--3801. IEEE, 2016.

\bibitem[Kufa and Kratochvil(2017)]{kufa2017software}
Jan Kufa and Tomas Kratochvil.
\newblock Software and hardware hevc encoding.
\newblock In \emph{2017 International Conference on Systems, Signals and Image Processing (IWSSIP)}, pages 1--5. IEEE, 2017.

\bibitem[Li et~al.(2024)Li, Ankireddy, Zhao, Nourkhiz~Mahjoub, Moradi~Pari, Topcu, Chinchali, and Kim]{li2024task}
Po-han Li, Sravan~Kumar Ankireddy, Ruihan~Philip Zhao, Hossein Nourkhiz~Mahjoub, Ehsan Moradi~Pari, Ufuk Topcu, Sandeep Chinchali, and Hyeji Kim.
\newblock Task-aware distributed source coding under dynamic bandwidth.
\newblock \emph{Advances in Neural Information Processing Systems}, 36, 2024.

\bibitem[Li et~al.(2021)Li, Shi, and Chen]{li2021task}
Xin Li, Jun Shi, and Zhibo Chen.
\newblock Task-driven semantic coding via reinforcement learning.
\newblock \emph{IEEE Transactions on Image Processing}, 30:\penalty0 6307--6320, 2021.

\bibitem[Liu et~al.(2016)Liu, Anguelov, Erhan, Szegedy, Reed, Fu, and Berg]{liu2016ssd}
Wei Liu, Dragomir Anguelov, Dumitru Erhan, Christian Szegedy, Scott Reed, Cheng-Yang Fu, and Alexander~C Berg.
\newblock Ssd: Single shot multibox detector.
\newblock In \emph{Computer Vision--ECCV 2016: 14th European Conference, Amsterdam, The Netherlands, October 11--14, 2016, Proceedings, Part I 14}, pages 21--37. Springer, 2016.

\bibitem[Liu et~al.(2008)Liu, Li, and Soh]{liu2008region}
Yang Liu, Zheng~Guo Li, and Yeng~Chai Soh.
\newblock Region-of-interest based resource allocation for conversational video communication of h. 264/avc.
\newblock \emph{IEEE transactions on circuits and systems for video technology}, 18\penalty0 (1):\penalty0 134--139, 2008.

\bibitem[Lou et~al.(2022)Lou, Lin, Marshall, Saupe, and Liu]{TranSalNet}
Jianxun Lou, Hanhe Lin, David Marshall, Dietmar Saupe, and Hantao Liu.
\newblock Transalnet: Towards perceptually relevant visual saliency prediction.
\newblock \emph{Neurocomputing}, 2022.

\bibitem[Lu et~al.(2019)Lu, Ouyang, Xu, Zhang, Cai, and Gao]{lu2019dvc}
Guo Lu, Wanli Ouyang, Dong Xu, Xiaoyun Zhang, Chunlei Cai, and Zhiyong Gao.
\newblock Dvc: An end-to-end deep video compression framework.
\newblock In \emph{Proceedings of the IEEE/CVF conference on computer vision and pattern recognition}, pages 11006--11015, 2019.

\bibitem[Mandhane et~al.(2022)Mandhane, Zhernov, Rauh, Gu, Wang, Xue, Shang, Pang, Claus, Chiang, et~al.]{mandhane2022muzero}
Amol Mandhane, Anton Zhernov, Maribeth Rauh, Chenjie Gu, Miaosen Wang, Flora Xue, Wendy Shang, Derek Pang, Rene Claus, Ching-Han Chiang, et~al.
\newblock Muzero with self-competition for rate control in vp9 video compression.
\newblock \emph{arXiv preprint arXiv:2202.06626}, 2022.

\bibitem[Mao et~al.(2020)Mao, Gu, Wang, Chen, Lazic, Levine, Pang, Claus, Hechtman, Chiang, et~al.]{mao2020neural}
Hongzi Mao, Chenjie Gu, Miaosen Wang, Angie Chen, Nevena Lazic, Nir Levine, Derek Pang, Rene Claus, Marisabel Hechtman, Ching-Han Chiang, et~al.
\newblock Neural rate control for video encoding using imitation learning.
\newblock \emph{arXiv preprint arXiv:2012.05339}, 2020.

\bibitem[Merritt and Vanam(2006)]{merritt2006x264}
Loren Merritt and Rahul Vanam.
\newblock x264: A high performance h. 264/avc encoder.
\newblock \emph{online] http://neuron2. net/library/avc/overview\_x264\_v8\_5. pdf}, 2006.

\bibitem[Muhammad and Yeasin(2020)]{muhammad2020eigen}
Mohammed~Bany Muhammad and Mohammed Yeasin.
\newblock Eigen-cam: Class activation map using principal components.
\newblock In \emph{2020 international joint conference on neural networks (IJCNN)}, pages 1--7. IEEE, 2020.

\bibitem[PIERROT et~al.(2021)PIERROT, Mac{\'e}, Sevestre, Monier, Laterre, Perrin, Beguir, and Sigaud]{pierrot2021factored}
Thomas PIERROT, Valentin Mac{\'e}, Jean-Baptiste Sevestre, Louis Monier, Alexandre Laterre, Nicolas Perrin, Karim Beguir, and Olivier Sigaud.
\newblock Factored action spaces in deep reinforcement learning.
\newblock 2021.

\bibitem[Pont-Tuset et~al.(2017)Pont-Tuset, Perazzi, Caelles, Arbel{\'a}ez, Sorkine-Hornung, and Van~Gool]{pont20172017}
Jordi Pont-Tuset, Federico Perazzi, Sergi Caelles, Pablo Arbel{\'a}ez, Alex Sorkine-Hornung, and Luc Van~Gool.
\newblock The 2017 davis challenge on video object segmentation.
\newblock \emph{arXiv preprint arXiv:1704.00675}, 2017.

\bibitem[Raffin et~al.(2021)Raffin, Hill, Gleave, Kanervisto, Ernestus, and Dormann]{raffin2021stable}
Antonin Raffin, Ashley Hill, Adam Gleave, Anssi Kanervisto, Maximilian Ernestus, and Noah Dormann.
\newblock Stable-baselines3: Reliable reinforcement learning implementations.
\newblock \emph{Journal of Machine Learning Research}, 22\penalty0 (268):\penalty0 1--8, 2021.

\bibitem[Schulman et~al.(2017)Schulman, Wolski, Dhariwal, Radford, and Klimov]{schulman2017proximal}
John Schulman, Filip Wolski, Prafulla Dhariwal, Alec Radford, and Oleg Klimov.
\newblock Proximal policy optimization algorithms.
\newblock \emph{arXiv preprint arXiv:1707.06347}, 2017.

\bibitem[Shi and Chen(2020)]{shi2020reinforced}
Jun Shi and Zhibo Chen.
\newblock Reinforced bit allocation under task-driven semantic distortion metrics.
\newblock In \emph{2020 IEEE international symposium on circuits and systems (ISCAS)}, pages 1--5. IEEE, 2020.

\bibitem[Shor and Johnston(2022)]{shor2022need}
Joel Shor and Nick Johnston.
\newblock The need for medically aware video compression in gastroenterology.
\newblock \emph{arXiv preprint arXiv:2211.01472}, 2022.

\bibitem[Singh et~al.(2022)Singh, Delp, and Reibman]{singh2022video}
Praneet Singh, Edward~J Delp, and Amy~R Reibman.
\newblock Video-analytics task-aware quad-tree partitioning and quantization for hevc.
\newblock In \emph{2022 IEEE International Conference on Image Processing (ICIP)}, pages 2936--2940. IEEE, 2022.

\bibitem[Sullivan et~al.(2012)Sullivan, Ohm, Han, and Wiegand]{sullivan2012overview}
Gary~J Sullivan, Jens-Rainer Ohm, Woo-Jin Han, and Thomas Wiegand.
\newblock Overview of the high efficiency video coding (hevc) standard.
\newblock \emph{IEEE Transactions on circuits and systems for video technology}, 22\penalty0 (12):\penalty0 1649--1668, 2012.

\bibitem[Sutton and Barto(1998)]{sutton1998introduction}
Richard~S. Sutton and Andrew~G. Barto.
\newblock \emph{Reinforcement Learning: {A}n Introduction}.
\newblock The MIT Press, Cambridge, MA, 1998.

\bibitem[Tomar(2006)]{tomar2006converting}
Suramya Tomar.
\newblock Converting video formats with ffmpeg.
\newblock \emph{Linux Journal}, 2006\penalty0 (146):\penalty0 10, 2006.

\bibitem[Van~Hasselt and Wiering(2007)]{van2007reinforcement}
Hado Van~Hasselt and Marco~A Wiering.
\newblock Reinforcement learning in continuous action spaces.
\newblock In \emph{2007 IEEE International Symposium on Approximate Dynamic Programming and Reinforcement Learning}, pages 272--279. IEEE, 2007.

\bibitem[Wang et~al.(2004)Wang, Bovik, Sheikh, and Simoncelli]{wang2004image}
Zhou Wang, Alan~C Bovik, Hamid~R Sheikh, and Eero~P Simoncelli.
\newblock Image quality assessment: from error visibility to structural similarity.
\newblock \emph{IEEE transactions on image processing}, 13\penalty0 (4):\penalty0 600--612, 2004.

\bibitem[Wenger(2003)]{wenger2003h}
Stephan Wenger.
\newblock H. 264/avc over ip.
\newblock \emph{IEEE transactions on circuits and systems for video technology}, 13\penalty0 (7):\penalty0 645--656, 2003.

\bibitem[Wiegand et~al.(2003)Wiegand, Sullivan, Bjontegaard, and Luthra]{wiegand2003overview}
Thomas Wiegand, Gary~J Sullivan, Gisle Bjontegaard, and Ajay Luthra.
\newblock Overview of the h. 264/avc video coding standard.
\newblock \emph{IEEE Transactions on circuits and systems for video technology}, 13\penalty0 (7):\penalty0 560--576, 2003.

\bibitem[Windsheimer et~al.(2024)Windsheimer, Brand, and Kaup]{windsheimer2024annotation}
Marc Windsheimer, Fabian Brand, and Andr{\'e} Kaup.
\newblock On annotation-free optimization of video coding for machines.
\newblock \emph{arXiv preprint arXiv:2406.07938}, 2024.

\bibitem[Wu et~al.(2024)Wu, Quan, He, Lai, Li, Yu, Lin, and Yang]{wu2024qs}
Chang Wu, Guancheng Quan, Gang He, Xin-Quan Lai, Yunsong Li, Wenxin Yu, Xianmeng Lin, and Cheng Yang.
\newblock Qs-nerv: Real-time quality-scalable decoding with neural representation for videos.
\newblock In \emph{ACM Multimedia 2024}, 2024.

\bibitem[Xie et~al.(2022)Xie, Li, Lin, Chen, Zhang, Zhang, and Li]{xie2022hierarchical}
Guangqi Xie, Xin Li, Shiqi Lin, Zhibo Chen, Li Zhang, Kai Zhang, and Yue Li.
\newblock Hierarchical reinforcement learning based video semantic coding for segmentation.
\newblock In \emph{2022 IEEE International Conference on Visual Communications and Image Processing (VCIP)}, pages 1--5. IEEE, 2022.

\bibitem[Ye et~al.(2021)Ye, Liu, Kurutach, Abbeel, and Gao]{ye2021mastering}
Weirui Ye, Shaohuai Liu, Thanard Kurutach, Pieter Abbeel, and Yang Gao.
\newblock Mastering atari games with limited data.
\newblock \emph{Advances in neural information processing systems}, 34:\penalty0 25476--25488, 2021.

\bibitem[Yu et~al.(2020)Yu, Chen, Wang, Xian, Chen, Liu, Madhavan, and Darrell]{yu2020bdd100k}
Fisher Yu, Haofeng Chen, Xin Wang, Wenqi Xian, Yingying Chen, Fangchen Liu, Vashisht Madhavan, and Trevor Darrell.
\newblock Bdd100k: A diverse driving dataset for heterogeneous multitask learning.
\newblock In \emph{Proceedings of the IEEE/CVF conference on computer vision and pattern recognition}, pages 2636--2645, 2020.

\bibitem[Zhang et~al.(2024)Zhang, Ahonen, Le, Yang, and Cricri]{zhang2024competitive}
Honglei Zhang, Jukka~I Ahonen, Nam Le, Ruiying Yang, and Francesco Cricri.
\newblock Competitive learning for achieving content-specific filters in video coding for machines.
\newblock \emph{arXiv preprint arXiv:2406.12367}, 2024.

\bibitem[Zhang et~al.(2022)Zhang, Sun, Jiang, Yu, Weng, Yuan, Luo, Liu, and Wang]{zhang2022bytetrack}
Yifu Zhang, Peize Sun, Yi Jiang, Dongdong Yu, Fucheng Weng, Zehuan Yuan, Ping Luo, Wenyu Liu, and Xinggang Wang.
\newblock Bytetrack: Multi-object tracking by associating every detection box.
\newblock In \emph{European conference on computer vision}, pages 1--21. Springer, 2022.

\end{thebibliography}
}
\clearpage
\setcounter{page}{1}
\newpage
\onecolumn

\section{Appendix}
\subsection{Environment and agent details}
\subsubsection{Environment details}
\label{sec:apx_env_details}
\begin{table}[htb!]
\centering
\begin{tabular}{|c|} 
 \hline
 \textbf{Global encoder statistics used as state information}\\ [0.5ex] 
 \hline\hline
 Next frame x264 selected QP value \\ 
 \hline
 Next frame number \\
 \hline
 Current bitstream size \\
 \hline
 Current frame x264 selected QP value \\
 \hline
 Average QP \\
 \hline
 Percentages of I type Macro Blocks \\
 \hline
 Percentages of P type Macro Blocks \\
 \hline
 Percentages of skip-type Macro Blocks \\
 \hline
 x264 calculated PSNR \\
 \hline
 x264 calculated SSIM \\
 \hline
 Percentages of bits used for Motion Vectors \\
 \hline
 Percentages of bits used for DCT coefficient \\
 \hline
 Progress of encoding \\
 \hline
  bit-rate error \\
 \hline
 Next frame type \\
 \hline
 Next frame complexity \\
 \hline
\end{tabular}
\caption{Detailed components of global encoder statistic used in state information}
\end{table}

\begin{table*}[htb!]
\centering
\begin{tabular}{|c|} 
 \hline
 \textbf{Local (per-MB) encoder statistic used as state information}\\ [0.5ex] 
 \hline\hline
 x264 energy values per Macro Block\\ [0.5ex] 
 \hline
 x264 intra encoding cost per Macro Block \\ 
 \hline
 x264 propagating encoding cost per Macro Block \\
 \hline
x264 inverse quantization scale factor per Macro Block  \\
 \hline
\end{tabular}
\caption{Detailed components of per-MB encoder statistic used in state information}
\end{table*}

\newpage
\subsubsection{Agent architecture} \label{appendix:architecture}
\begin{figure*}[h]
    \centering
    \includegraphics[width=1.\textwidth]{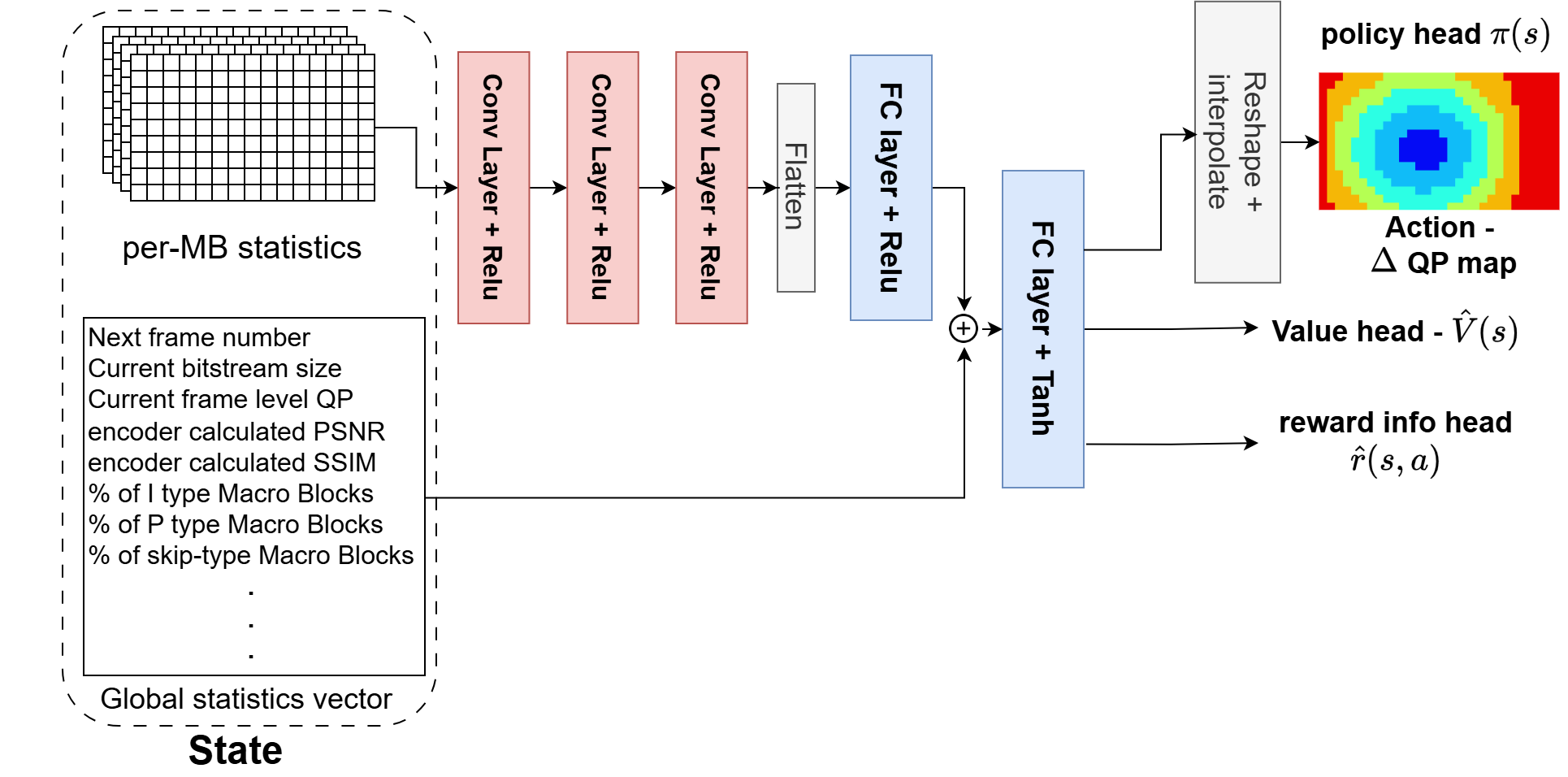}
    \caption{\ourmethod{} agent architecture; Input is the statistics from the encoder, the output is the delta QP map}
    \label{fig:agent_arch}
\end{figure*}

To train the policy, we use the PPO algorithm \citep{schulman2017proximal}, where the architecture of the policy is as follows: The per-block statistics are processed through a compact convolutional neural network (CNN) comprising three convolutional layers. These layers employ kernel sizes of $3x3$ or $4x4$ with a stride of $1$. The resulting features are subsequently flattened and concatenated with the global statistics. A fully connected layer then derives a latent representation of dimension $64$. This latent representation serves as input to three distinct fully connected networks: the value network (critic), the policy network (actor), and the reward prediction network described in the following subsection. A diagram of the full system is given in Figure~\ref{fig:agent_arch}. The agent's stochastic policy is modeled using a diagonal multivariate Gaussian distribution, where the agent learns the state-dependent mean vectors while maintaining independent standard deviation parameters for each dimension.

\newpage{}
\subsection{More details on RD-curves and BD-rate}
The effectiveness of video compression is typically measured by comparing the compressed video's file size and visual quality to the original. Metrics like Peak Signal-to-Noise Ratio (PSNR) and Structural Similarity Index Measure (SSIM;~\cite{wang2004image}) are often used to objectively assess quality. When comparing two encoders the compression efficiency is usually considered. To do so, a video is encoded in several desired bit-rates with each encoder to form a rate-distortion (RD) curve, where the $y$ axis is the quality measure, e.g. PSNR. 
If one encoder's curve is shifted-left than the other, it means it requires less bits to reach the same quality, rendering it more efficient. If we integrate over the entire curve, and average the result over multiple videos, we obtain a quantity specifying how much bits saves one encoder than the other, a quantity referred to as Bjontegaard delta rate (BD-rate) \citep{wiegand2003overview}. 

With the increasing usages of videos for machine vision, many researchers have recognized the need for task-aware compression and proposed a suitable evaluation metric \citep{kong2016new, shi2020reinforced}. The most straightforward metric which we also use in this paper is obtained by replacing the PSNR in the RD-curve (the $y$-axis) with a task-specific loss measure such as mIOU or detection precision. 

\newpage{}
\subsection{Further results}
\textbf{Full results tables with bit-rate error:}
Here, we provide all of the tables used in the main text with the bit-rate error data.
\begin{table}[h]
    \centering
    \begin{tabular}{lccc}
    \toprule 
    \textbf{ROI encoding} & \textbf{Saliency-weighted} & \textbf{PSNR} & \textbf{Bit-rate} \\
    \textbf{experiment} & \textbf{PSNR BD-rate} & \textbf{BD-rate} & \textbf{error $[1e-3]$} \\
    \midrule
    \ourmethod{} & $-25.64 \pm 0.99$ & $-5.26 \pm 0.36$ & $-1.0 \pm 0.43$ \\
    \bottomrule
    \end{tabular}
    \caption{Results on \ourmethod{} applied on the test-set for the saliency-weighted PSNR task.}
    \label{tab_apx:saliencyw}
\end{table}

\begin{table}[h]
    \centering
    \begin{tabular}{lcccc}
    \toprule
    \textbf{} & \textbf{Precision} & \textbf{Recall} &\textbf{PSNR} &   \textbf{Precision}  \\
     \textbf{} & \textbf{(YOLO)} & \textbf{(YOLO)} & &   \textbf{(SSD)}\\
    \midrule
    \ourmethod{}& $-24.7 \pm 1.57$ & $-19.75 \pm 2.97$ & $1.19 \pm 0.46$ &  $-26.2 \pm 1.48$\\ 
    \toprule
     \textbf{} &  \textbf{Recall} & \textbf{Segmentation} & \textbf{Bit-rate} & \\
     \textbf{} &  \textbf{(SSD)} &  \textbf{IOU} & \textbf{error} $[1e-3]$ & \\
         \midrule
     \ourmethod{}& $-25.81 \pm 2.03$ & $-14.6 \pm 1.81$ & $0.13 \pm 0.44$ &\\ 
    \bottomrule
    \end{tabular}
        \caption{
        BD-rate Results on \ourmethod{} applied on test set for the car detection task for various settings. Negative values mean that \ourmethod{} improves over baseline.}
    \label{tab_apx:car-detection}
\end{table}

\begin{table}[h]
    \centering
    \begin{tabular}{lcccc}
    \toprule 
    \textbf{} & \textbf{Precision} & \textbf{Recall} & \textbf{PSNR} & \textbf{Bit-rate}\\
    \textbf{Car detection} & \textbf{BD-rate}& \textbf{BD-rate} & \textbf{BD-rate} & \textbf{error $[1e-3]$} \\
    \midrule
    \ourmethod{} & $-24.7 \pm 1.57$ & $-19.75 \pm 2.97$  & $1.19 \pm 0.46$ & $0.13 \pm 0.44$ \\
    \ourmethod{} w/o RI & $-19.4 \pm 1.38$ & $-11.94 \pm 1.7$ & $1.92 \pm 0.41$ & $0.4 \pm 0.47$ \\
    \ourmethod{} $\gamma=0$ & $-9.78 \pm 1.29$ & $-10.28 \pm 2.14$  & $5.44 \pm 0.6$ & $2.4 \pm 0.44$ \\
    \midrule
    \textbf{} & \textbf{Saliency-weighted} & \textbf{PSNR} & \textbf{Bit-rate} &\\
    \textbf{ROI encoding} & \textbf{PSNR BD-rate} & \textbf{BD-rate} & \textbf{error $[1e-3]$} &\\
    \midrule
    \ourmethod{} & $-25.64 \pm 0.99$ & $-5.26 \pm 0.36$ & $-1.0 \pm 0.43$ &\\
    \ourmethod{} w/o RI & $-23.46 \pm 0.97$ & $-4.54 \pm 0.42$ & $5.3 \pm 0.48$ &\\
    \ourmethod{} $\gamma=0$ & $-16.01 \pm 0.77$ & $2.11 \pm 0.31$ & $6.9 \pm 0.43$ &\\
    \bottomrule
    \end{tabular}
    \caption{Ablation study. (1) Full \ourmethod{} (2) Omitting reward information (RI) from the training process and (3) Ignoring long term effects by using a myopic policy.} 
    \label{tab_apx:ablation}
\end{table}

\begin{figure}[ht]
    \centering
    \includegraphics[width=0.9\linewidth]{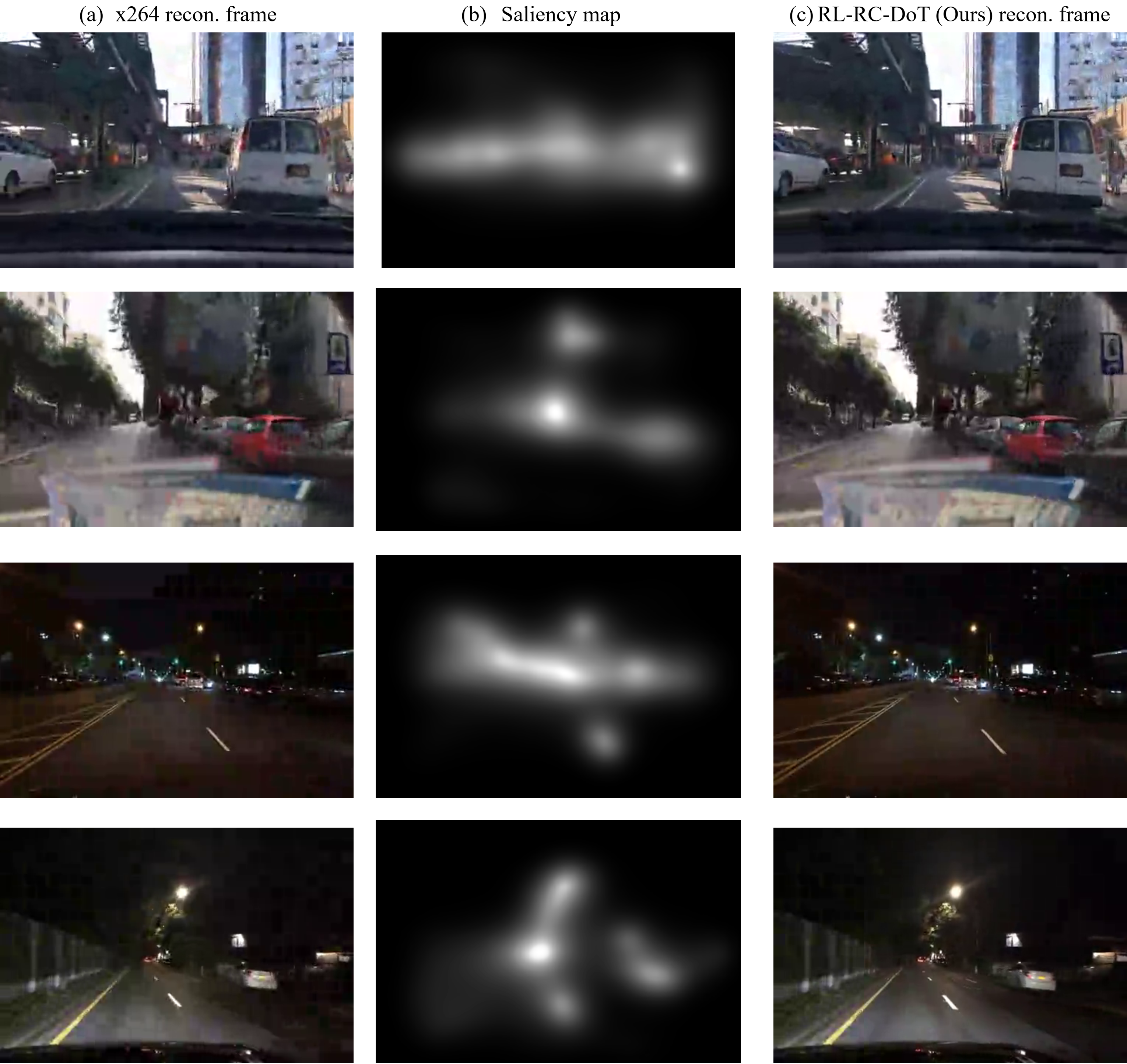}
    \caption{Saliency weighted PSNR results. (a) x264 reconstructed frame, (b) Saliency map of raw frame, extracted with ~\citep{TranSalNet} (c) \ourmethod{} reconstructed frame. Notice that both \ourmethod{} and x264 used the same target bit-rate}
    \label{fig:more_sal}
\end{figure}

Here we give more qualitative results, Figures \ref{fig:more_sal} and \ref{fig:more_sal} gives more detection comparison between \ourmethod{} and x264.
\begin{figure}[ht]
    \centering
    \includegraphics[width=0.9\linewidth]{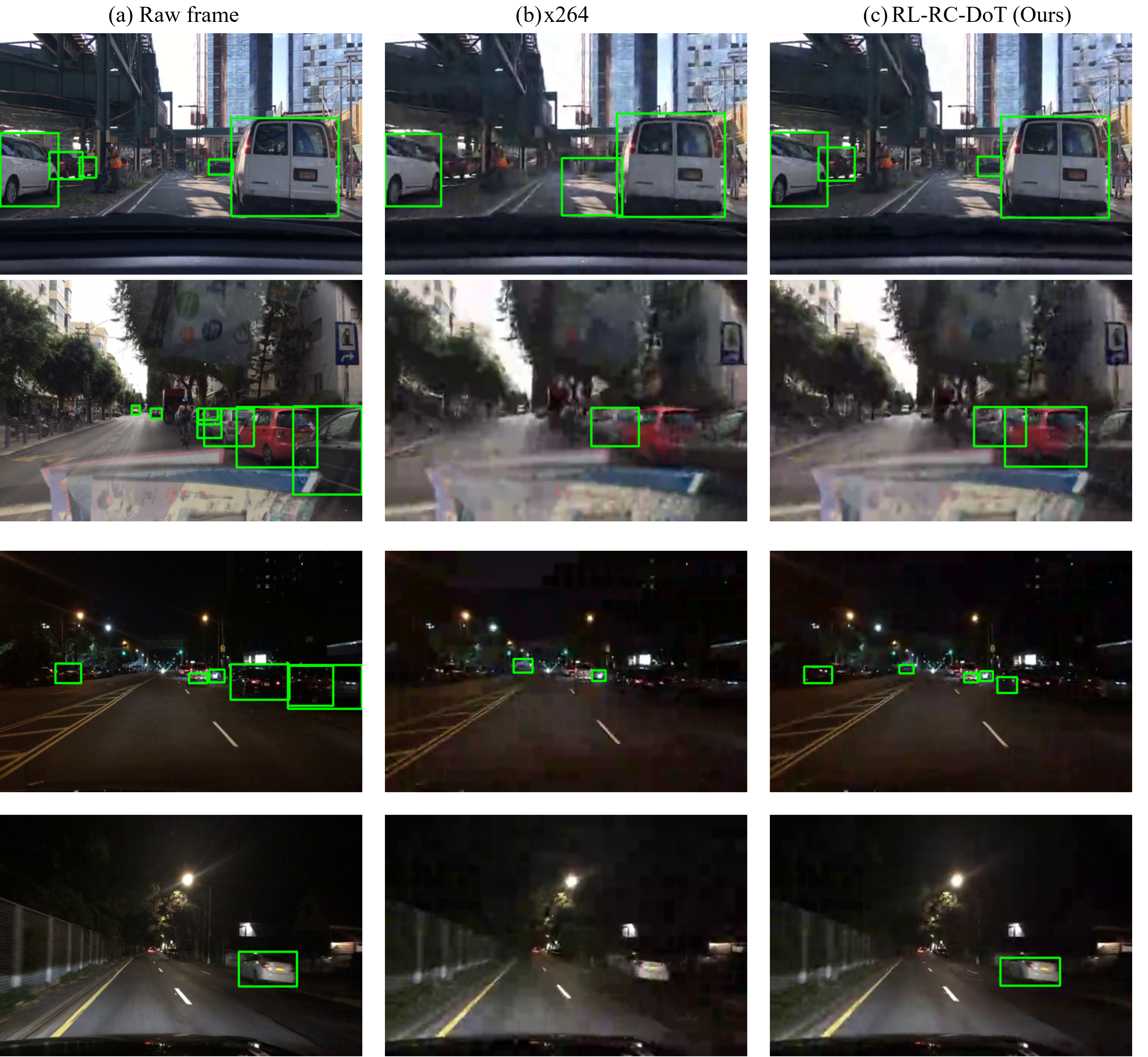}
    \caption{Car detection example result. (a) detection output on x264 reconstructed frame, (b) output on raw frame and (c) output on \ourmethod{} reconstructed frame. Notice that both \ourmethod{} and x264 used the same target bit-rate}
    \label{fig:more_detection}
\end{figure}

\begin{figure}[ht]
    \centering
    \includegraphics[width=0.9\linewidth]{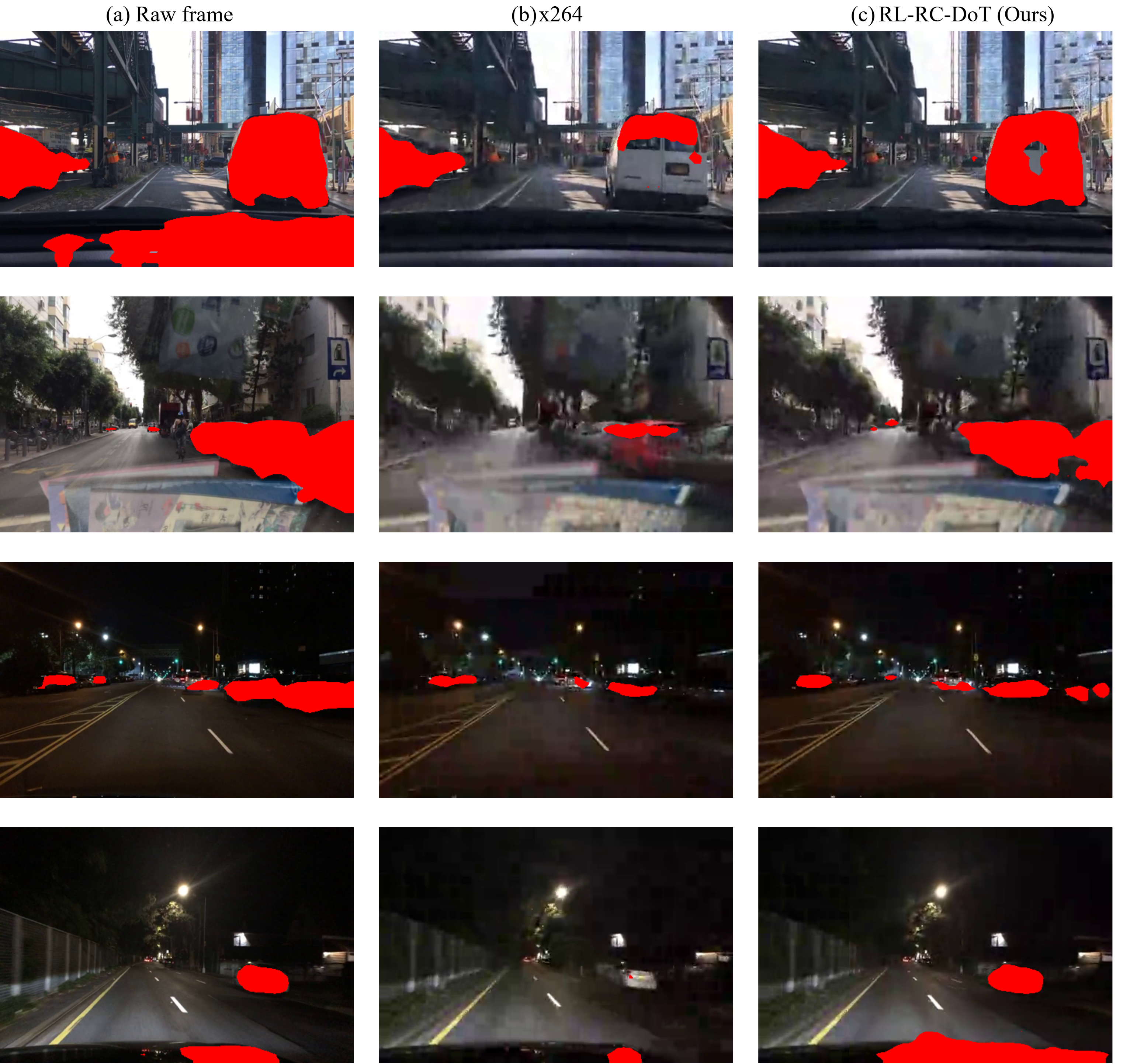}
    \caption{Car segmentation result comparison. (a) segmentation output on x264 reconstructed frame, (b) output on raw frame and (c) output on \ourmethod{} reconstructed frame. Notice that both \ourmethod{} and x264 used the same target bit-rate}
    \label{fig:model_segmentation}
\end{figure}

\newpage
\textbf{Qualitative results on task-robustness:}
Figure \ref{fig:model_segmentation}, shows a qualitative example of task robustness. We compare the images in both types of rate-control, and the output of the downstream task. We can see the details corresponding to the downstream task are better reconstructed yielding a more relevant image.


\subsubsection{Additional downstream tasks}
We evaluated \ourmethod{} with two more downstream tasks. First, in \textbf{video segmentation} using the \href{davischallenge.org/}{DAVIS} dataset \citep{pont20172017}. \ourmethod{} reduced BD-rate by $8\%$ over x264. Quality was measured by IoU of a Mask R-CNN \cite{he2017mask} segmentation. Second, in a \textbf{tracking} task, where the model was pre-trained for detection using the BDD dataset and tested in a different task of multi-object tracking (task shift). \ourmethod{} reduced BD-rate by $3.2\%$ over x264. Quality was measured by MOTA using ByteTracker \cite{zhang2022bytetrack}.

\newpage
\subsubsection{Different x264 configurations}
As mentioned in section ~\ref{subsec:exp_details}. We only trained our agent on the \textit{medium} preset configuration of x264. To test \ourmethod{} generalization palpabilities, we tested the agent on different preset configurations of x264. The BD-rate saving for car detecion precision are shown in table \ref{tab:presets}.
\begin{table}
    \centering
    \begin{tabular}{l|ccccccc}
    \toprule
    x264 preset & superfast & fast & veryfast & medium & slower & slow & veryslow\\
    \midrule
    Precision BD-rate & -14.5 & -22/8 & -21.9 & -24.7 & -23.2 & -25.2 & -24.9\\
    \bottomrule
    \end{tabular}
    \caption{Test BD-rate reduction of RL-RC-DoT for various x264 presets on car detection. Negative values mean that RL-RC-DoT improves over x264.}
    \label{tab:presets}
\end{table}

\newpage
\subsubsection{Quantization map analysis}
\begin{figure}[ht]
    \centering
    \includegraphics[width=1\linewidth]{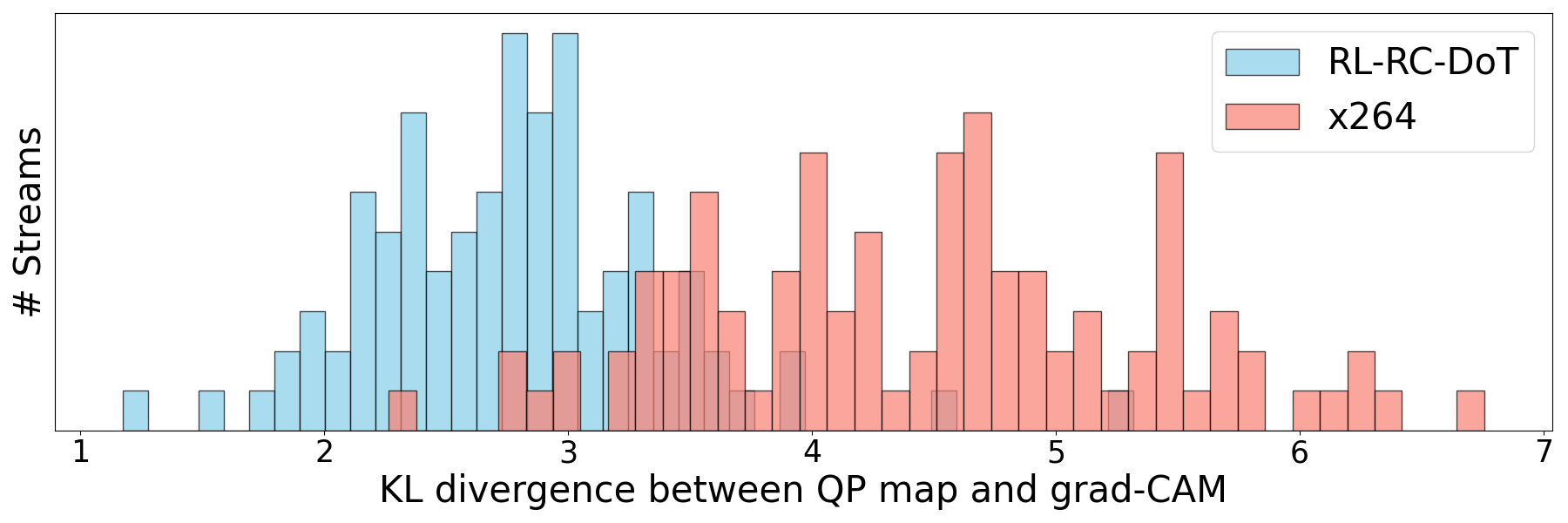}
    \vspace{-18pt} 
    \caption{Distribution of average KL divergence between QP maps and Eigen-CAM.}
    \label{fig:qpmaps-stat}
    \vspace{-6mm}
\end{figure}

To gain more insight into how our approach affects QP map, we quantified the relation between the QP map and areas in the video that are useful for car detection. Specifically, we computed the KL-divergence between normalized QP-maps and \href{https://arxiv.org/abs/2008.00299}{Eigen-CAM} \cite{muhammad2020eigen} spatial map. Figure \ref{fig:qp_maps} illustrates the three maps for one frame, showing that RL-RC-DoT allocates more bits to areas in the frame that are informative for car detection.  
Figure \ref{fig:qpmaps-stat} quantifies this effect across our full test data (mean $D_{kl}$ RL-RC-DoT = $2.6$ 
, x264=$4.4$).


\newpage
\subsection{Task accuracy to distortion trade-off}
\label{sec:accuracy_psnr_tradeoff}
As previously discussed, \ourmethod{} gains BD-rate reductions of $24.7\% (\pm 1.38\%)$ with respect to car detection precision task, while paying a minimal cost to overall video quality, as evidenced by a slight increase in PSNR BD-rate of $1.19\% (\pm 0.46\%)$. This is important since we want video to still be watchable by human eyes, for validation purposes and robustness to changing task models.

\begin{figure}[ht]
    \centering
    \includegraphics[width=0.6\linewidth]{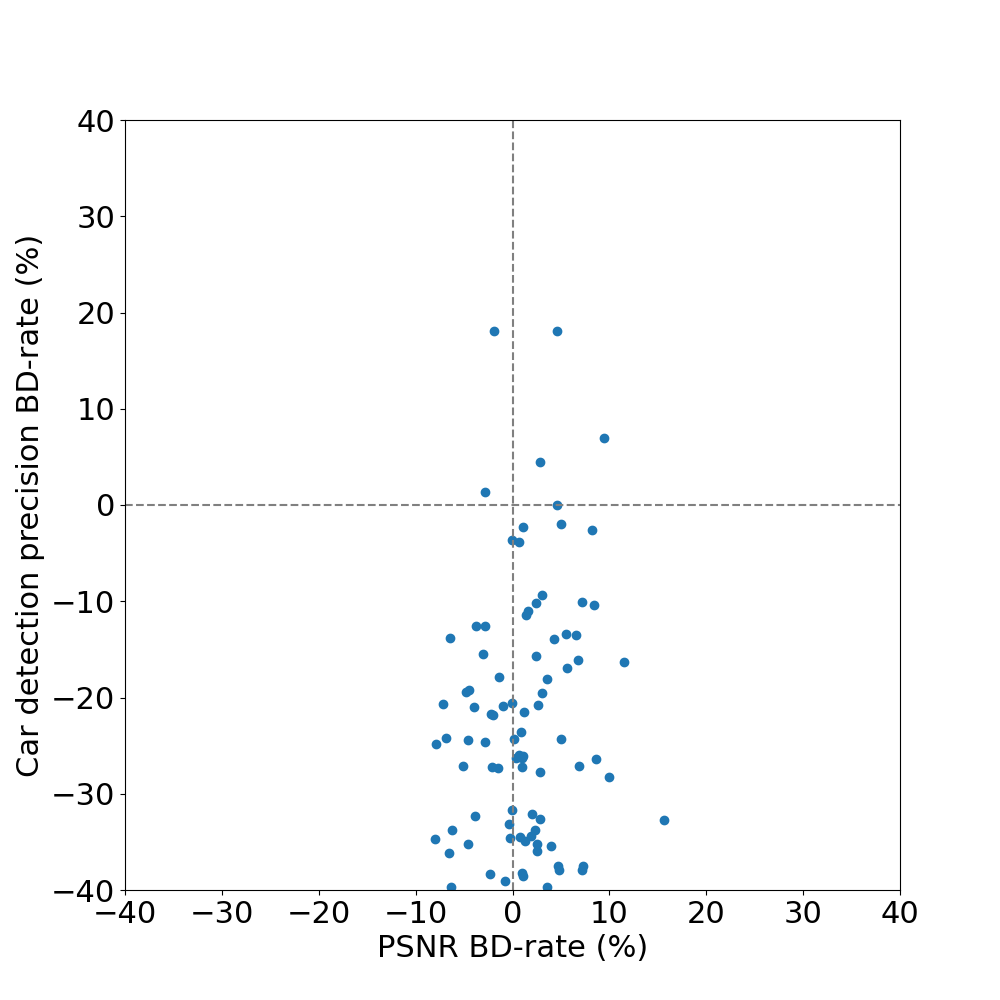}
    \caption{PSNR BD-rate to detection precision BD-rate, where each point represent a single stream in the test set}
    \label{fig:psnr_2_precision}
\end{figure}

To further illustrate this point, in Figure \ref{fig:psnr_2_precision} we show the PSNR and task performance BD-rate obtained by \ourmethod{} for each stream in the test set. In the plots we see the PSNR varies around $0$ while the tasks performance is well below.

\newpage
\subsection{Action Space Resolution}\label{subsec:mbmult}

\begin{figure}[h]
     \centering
     \includegraphics[width=0.6\textwidth]{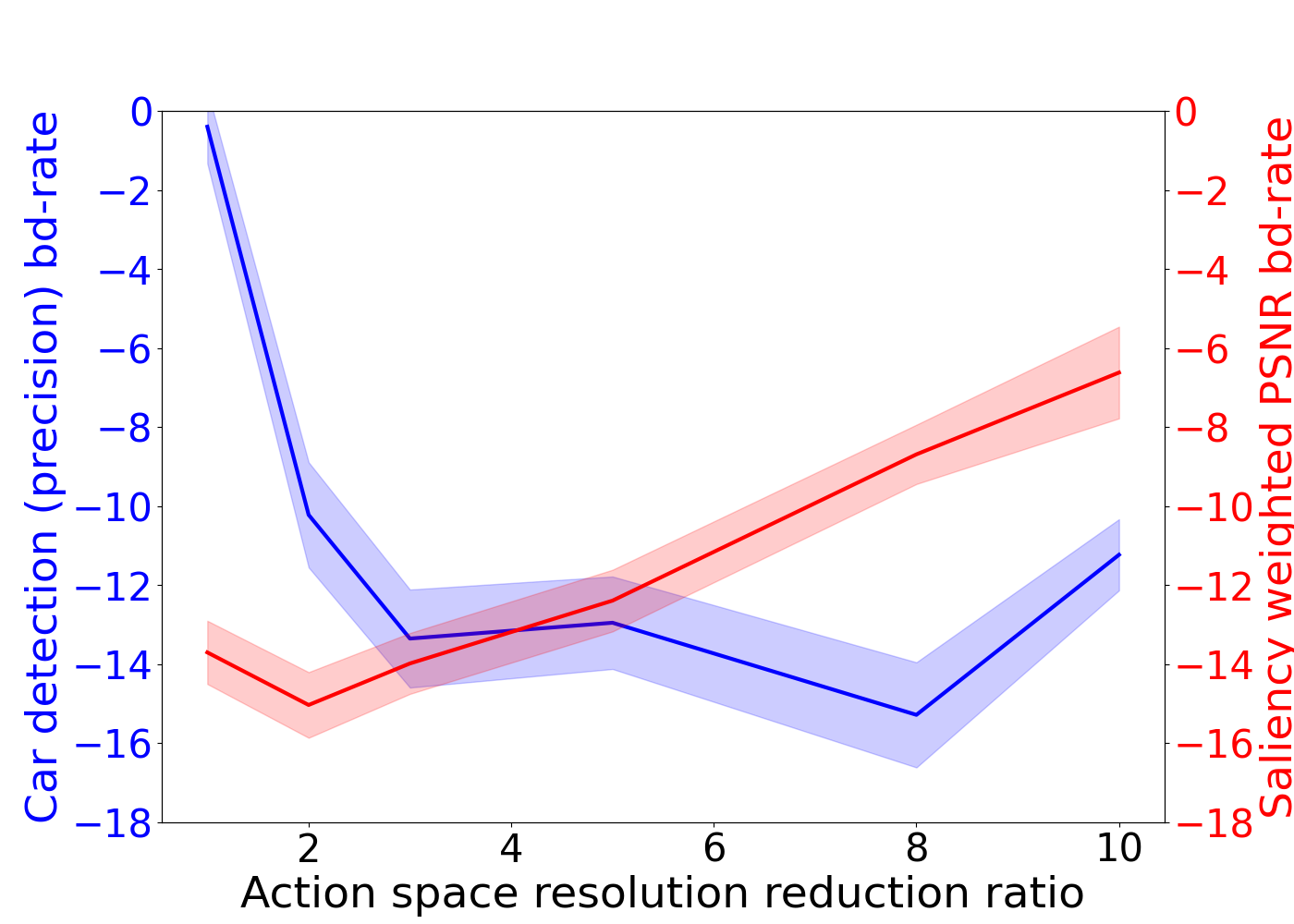}     
     \caption{The effect of action space resolution on the BD-rate for both tasks}
     \label{fig:mb_mult}
\end{figure}

Since we show our results on a videos of size 480x320 with macro-blocks of size 16x16, the action space is of size 30x20. The size of the action space drastically affects the performance of the agent and the convergence rate of the training process. Thus, we propose to set a lower resolution action space and upsample to the original action space by interpolation. The trade-off here is clear -- if we make decisions in high resolution, the agent can take a long time to converge, whereas a low resolution decision will not provide the finer control required for accurate bit allocation for the downstream-task resulting in a sub-optimal performance. We illustrate this notion in Figure \ref{fig:mb_mult}. We plot the task BD-rate for multiple choices of resolution reduction ratios for each of the tasks. The plot indeed shows the trade-off between the two, where each task has a different optimal choice for action space resolution. We note that these results may depend on the number of frames allotted for training, where we expect longer training to benefit lower resolutions. 

\newpage
\subsection{BDD100K streams}
\label{sec:bdd100k}
Here we elaborate on the streams we used from bdd100k dataset  \citep{yu2020bdd100k}:
\begin{table*}[b]
\centering
\begin{tabular}{||c|c||} 
 \hline
 \multicolumn{2}{|c|}{\textbf{Train streams}} \\ [0.5ex] 
 \hline\hline
0000f77c-6257be58 & 000e0252-8523a4a9\\
\hline 
000f157f-dab3a407 & 000f8d37-d4c09a0f\\
\hline 
00a04f65-af2ab984 & 00a0f008-3c67908e\\
\hline 
00a0f008-a315437f & 00a1176f-0652080e\\
\hline 
00a1176f-5121b501 & 00a2e3ca-5c856cde\\
\hline 
00a2e3ca-62992459 & 00a2f5b6-d4217a96\\
\hline 
00a395fe-d60c0b47 & 00a9cd6b-b39be004\\
\hline 
00abd8a7-ecd6fc56 & 00abf44e-04004ca0\\
\hline 
00adbb3f-7757d4ea & 00afa5b2-c14a542f\\
\hline 
00afa6b9-4efe0141 & 00b04b30-501822fa\\
\hline 
00b1dfed-a89dbe2b & 00be7020-457a6db4\\
\hline 
00beeb02-ba0790aa & 00c12bd0-bb46e479\\
\hline 
00c29c52-f9524f1e & 00c41a61-4ba25ad4\\
\hline 
00c497ae-595d361b & 00c87627-b7f6f46c\\
\hline 
00ca8821-db8033d5 & 00cb28b9-08a22af7\\
\hline 
00ccf2e8-59a6bfc9 & 00ccf2e8-ac055be6\\
\hline 
00ccf2e8-f8c69860 & 00ce6f6d-50bbee62\\
\hline 
00ce8219-12c6d905 & 00ce8219-d0b5582e\\
\hline 
00cef86b-204ea619 & 00cef86b-d8d105b9\\
\hline 
00cf8e3d-3d27efb0 & 00cf8e3d-4683d983\\
\hline 
00cf8e3d-773de15e & 00cf8e3d-a7b4978c\\
\hline 
00d0f034-6d666f7b & 00d18b13-52d3e4c4\\
\hline 
00d4b6b7-7d0a60bf & 00d4b6b7-a0b1a3e0\\
\hline 
00d7268f-fd4487be & 00d79c0a-23bea078\\
\hline 
00d79c0a-a2b85ca4 & 00d84b1d-21e6fe01\\
\hline 
00d8944b-e157478b & 00d8d95a-74aa476a\\
\hline 
00d9e313-7d75bb18 & 00d9e313-926b6698\\
\hline 
00dc5030-237e7f71 & 00de601c-858a8a8d\\
\hline 
00de601c-cfa2404b & 00e49ed1-9d41220c\\
\hline 
00e4cae5-c0582574 & 00e5e793-f94de032\\
\hline 
00e81dcc-b1dd9e7b & 00e8c106-e197c4b1\\
\hline 
00c50078-6298b9c1 & 00b93c6e-6298aa25\\
\hline 
0000f77c-cb820c98 & \\
\hline
\end{tabular}
\caption{List of streams used in training}
\end{table*}

\begin{table*}[b]
\centering
\begin{tabular}{||c|c||} 
 \hline
 \multicolumn{2}{|c|}{\textbf{Validation streams}} \\ [0.5ex] 
 \hline\hline
00d8d95a-47d98291 & 00e02d60-54df99d1\\
\hline 
00a820ef-d655700e & 00ce95b0-84be34a3\\
\hline 
00d15d58-9197cde54 & 00b04b12-a7d7eb85\\
\hline 
00c17a92-d4803287 & \\
\hline 
\end{tabular}
\caption{List of streams used in validation}
\end{table*}

\begin{table*}
\centering
\begin{tabular}{||c|c|c|c||} 
 \hline
 \multicolumn{4}{|c|}{\textbf{Test streams}} \\ [0.5ex] 
 \hline\hline
cd35ea13-f49ee278 & cd389564-8be2128e & cdc05b0a-3bb83a9c & cd389564-9053f5fc\\
\hline
cd3b1173-63cb9e2e & cd3dab20-1b3e564e & cd3dab20-4ea3d971 & cd3df92f-d04e142c\\
\hline
cd40cb21-18170d03 & cd4ac25c-61a9eb11 & cd4bf816-2abb75c9 & cd4bf816-c2f9bf78\\
\hline
cd4ce4e5-6994fd2d & cd4ce4e5-d0968ec0 & cd4da443-da4fe8c7 & cd4deee2-0703d1c7\\
\hline
cd4deee2-1d9539bd & cd4deee2-37c8b95c & cd4deee2-3feadd6e & cd4deee2-60291439\\
\hline
cd4deee2-688c8bba & cd4deee2-8e12e5b5 & cd4deee2-9c9f6da1 & cd4deee2-adc7e92a\\
\hline
cd4deee2-ce4f69f5 & cd4deee2-d078d54a & cd547736-3b63cb96 & cd583365-462cca17\\
\hline
cd5a94cf-345f214a & cd5a9e1b-86faac85 & cd5b2540-465c9328 & cd5b2540-913cb8f7\\
\hline
cd5bee17-bef4f177 & cd5db4e0-1189ff83 & cd6af452-e54a1e36 & cd6c087e-03ca2127\\
\hline
cd6fdd33-ac9cb2db & cd704168-1231930e & cd7c12c7-7029da5d & cd7c12c7-9b46c2a8\\
\hline
cd7c92a7-3b20257f & cd7c92a7-89b23268 & cd7c92a7-9222ee19 & cd7c92a7-ed0d3926\\
\hline
cd7ee0b1-dd286a1b & cd7fb8f1-3d347a66 & cd828461-db8b4612 & cd839842-cd859db0\\
\hline
cd8b00aa-4aac0701 & cd8b00aa-5c017145 & cd8b00aa-f00ad3b9 & cd8b30b0-51369077\\
\hline
cd8b30b0-e8d12cc4 & cd8d2fde-2d2a3211 & cd9b6b86-9f62a970 & cd9b6b86-be582832\\
\hline
cd9cd3dd-d67bf5b6 & cd9d84d4-f59d3feb & cd9dff27-94731aba & cd9e7e2b-4b274850\\
\hline
cda33556-28510da1 & cda33556-8dc294b4 & cda33556-c6b3dd45 & cda55704-362ddfea\\
\hline
cda55704-754aac99 & cda63e8d-0afbf52b & cda63e8d-76b2fa43 & cda9acc1-1a92349d\\
\hline
cda9acc1-4469e473 & cda9acc1-9d1ef61a & cdac4037-afed765d & cdac7315-fe37a1d9\\
\hline
cdae6e60-0fb06a75 & cdae6e60-334ffc87 & cdae6e60-b729f2e6 & cdaee377-1eccb13a\\
\hline
cdaee377-2263611a & cdaee377-2b38ae2c & cdb06fa9-cfb70e11 & cdb06fa9-eba5643a\\
\hline
cdb3b01b-673f85b7 & cdb616df-393f382c & cdb688d4-33f24ca3 & cdb6b049-c96359c8\\
\hline
cdb815da-d03b9395 & cdb992be-f0f1613c & cdbb20a9-bdab1f4e & cdbbac37-49c0a335\\
\hline
cdbc7842-b72c4915 & cdbd1882-bdd416ea & cdbeedfd-4ab64af8 & cdbf4bd1-0c65ed7a\\
\hline
cdc05b0a-3bb83a9c & cdc05b0a-c53c36a6 & cdc05b0a-c6e8b6ec & cdc05b0a-ce908cf7\\
\hline
cdc05b0a-d4ff800b & cd3dab20-1b3e564e & cdc05b0a-efb78be5 & cdc05b0a-f2a67b44\\
\hline
\end{tabular}
\caption{List of streams used in test}
\end{table*}

\subsection{Reproducibility of experiments}

\textbf{Encoder environment: }To apply rate-control on the environment we changed the code of the open source x264 \citep{merritt2006x264} encoder so that in each frame it can obtain delta-QP values externally and provide relevant statistics as described in Appendix~\ref{sec:apx_env_details}. 

\textbf{RL Agent: }We provide a description of the policy's architecture in Appendix~\ref{appendix:architecture}. The agent was trained using PPO implementation from stable-baselines3 \citep{raffin2021stable} with default parameters, where we just added an MSE prediction loss (with weight $0.1$) for reward info. We used $\lambda=20$ to average between the bit-rate and downstream task rewards. 

\textbf{Experiments: } In our experiments we used the publicly available BDD100K dataset (\ref{subsec:dataset}) which was resized using the open source package ffmpeg\cite{tomar2006converting}. We provide the named list of streams we used in Appendix~\ref{sec:bdd100k}. In the experimental details subsection \ref{subsec:exp_details} we provide additional information on the hardware we used and the downstream task models we used for our experiments.

\textbf{Hardware:} To facilitate efficient training, we utilize 8 parallel environments running on an Intel(R) Xeon(R) CPU E5-2698 v4 @ 2.20GHz, complemented by an NVIDIA Tesla V100 32GB GPU. Each agent undergoes training on 20 million frames, a process that spans approximately 4 days.

\end{document}